\documentclass[10pt,twocolumn,letterpaper]{article}

\usepackage[pagenumbers]{cvpr} 

\usepackage{graphicx}
\usepackage{subcaption}
\usepackage{caption}
\usepackage[table,xcdraw]{xcolor}
\usepackage{makecell}
\usepackage{booktabs}
\usepackage{array}

\usepackage{float}
\usepackage{lipsum}
\usepackage{multirow}
\usepackage{pifont}
\usepackage{amsmath}
\usepackage{enumitem}

\usepackage{tcolorbox}
\usepackage[utf8]{inputenc}

\usepackage{bm}

\definecolor{cvprblue}{rgb}{0.21,0.49,0.74}
\usepackage[pagebackref,breaklinks,colorlinks,allcolors=cvprblue]{hyperref}

\def\paperID{*****} 
\def\confName{CVPR}
\def\confYear{2026}

\title{Unified Diffusion Transformer for High-fidelity Text-Aware Image Restoration}

\author{
Jin Hyeon Kim$^1$~\quad Paul Hyunbin Cho$^1$~\quad Claire Kim$^1$~\quad  \vspace{0.4em}\\ Jaewon Min$^1$~\quad Jaeeun Lee$^1$~\quad Jihye Park$^2$~\quad Yeji Choi$^1$~\quad Seungryong Kim$^1$ \\ \\
$^1$KAIST AI \qquad $^{2}$Samsung Electronics\\
{\tt \href{https://cvlab-kaist.github.io/UniT}{\textcolor{purple}{https://cvlab-kaist.github.io/UniT}}}
}

\begin{document}
\maketitle




\begin{abstract}
Text-Aware Image Restoration (TAIR) aims to recover high-quality images from low-quality inputs containing degraded textual content. While diffusion models provide strong generative priors for general image restoration, they often produce text hallucinations in text-centric tasks due to the absence of explicit linguistic knowledge. To address this, we propose \textbf{UniT}, a unified text restoration framework that integrates a Diffusion Transformer (DiT), a Vision-Language Model (VLM), and a Text Spotting Module (TSM) in an iterative fashion for high-fidelity text restoration. In UniT, the VLM extracts textual content from degraded images to provide explicit textual guidance. Simultaneously, the TSM, trained on diffusion features, generates intermediate OCR predictions at each denoising step, enabling the VLM to iteratively refine its guidance during the denoising process. Finally, the DiT backbone, leveraging its strong representational power, exploit these cues to recover fine-grained textual content while effectively suppressing text hallucinations. Experiments on the SA-Text and Real-Text benchmarks demonstrate that UniT faithfully reconstructs degraded text, substantially reduces hallucinations, and achieves state-of-the-art end-to-end F1-score performance in TAIR task. 
\end{abstract}

\section{Introduction}
\label{sec:1_intro}

\begin{figure}[t]
    \centering
    \includegraphics[width=\linewidth]{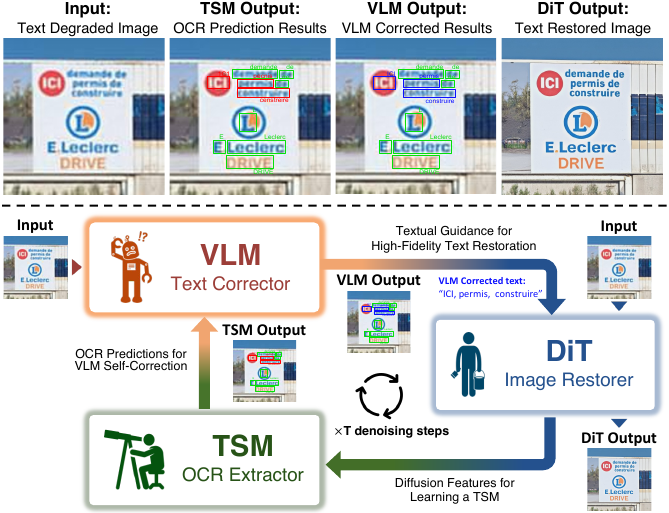}
    \vspace{-15pt}
    \caption{\textbf{UniT framework.} UniT consists of three components: a vision-language model (VLM), a text spotting module (TSM), and a diffusion transformer (DiT), each responsible for a distinct role in the text-aware image restoration task. The incorrect OCR prediction from TSM and the corrected results from the VLM are highlighted in {\color{red}red} and {\color{blue}blue}, respectively (best viewed with zoom).}
    \label{fig:unit_teaser}
    \vspace{-10pt}
\end{figure}
Image restoration~\cite{lin2024diffbir, wu2024seesr, duan2025dit4sr, liang2021swinir, wang2024exploiting, yue2023resshift, chen2024faithdiff} is a fundamental task in computer vision that aims to recover a high-quality (HQ) image from a low-quality (LQ) input. An important subtask is Text-Aware Image Restoration (TAIR)~\cite{min2025text, liu2023textdiff, noguchi2024scene, singh2024dcdm, ye2025textsr}, in which the LQ image contains degraded textual content, such as street signs, building logos, banners, and documents. Unlike general image restoration, where minor inaccuracies in texture and pattern recovery may be acceptable, even slight distortions in restored texts can hinder readability or alter the intended meaning. This makes TAIR a crucial task for real-world applications such as autonomous driving~\cite{tabernik2019deep, ertler2020mapillary}, AR/VR navigation~\cite{li2020textslam, strecker2022socrar}, and image enhancement~\cite{feng2023deep, feng2025docscanner}.

Leveraging the powerful generative priors of diffusion models~\cite{ho2020denoising, rombach2022high, podell2023sdxl, peebles2023scalable} has proven highly effective for various image restoration tasks~\cite{lin2024diffbir, wu2024seesr, duan2025dit4sr, liang2021swinir, wang2024exploiting, yue2023resshift, chen2024faithdiff}. Although these models can partially recover degraded textual content, a major challenge arises when they are applied to text-centric restoration. In such cases, diffusion models often produce text hallucinations, generating incorrect noisy symbols rather than accurately recovering the original text content (Fig.~\ref{fig:qual}). This limitation stems from the fact that diffusion models operate solely on visual priors, lacking explicit linguistic knowledge or character-level semantics. Without such priors, the model tends to generate strokes that appear plausible but do not correspond to valid text~\cite{min2025text}.

To address this issue, recent TAIR approaches introduce auxiliary text supervision to compensate for the diffusion model’s lack of linguistic cues. They incorporate text priors~\cite{ma2023text, ma2022text}, multimodal cues~\cite{zhao2022c3, singh2024dcdm}, recognition-based constraints~\cite{chen2022text, shi2016end}, or glyph-structure guidance~\cite{hu2025text, luo2025restore} to guide the text restoration process. Despite their effectiveness, these methods rely on complex multi-module architectures, require careful balancing of multiple objectives, and remain fundamentally constrained by the limited linguistic capacity of diffusion and OCR modules.

On one hand, Vision-Language Models (VLMs)~\cite{liu2023visual, Qwen2.5-VL, chen2024internvl, lu2024ovis} offer a powerful means of extracting textual information from LQ images by leveraging their rich visual and linguistic knowledge. In the context of text restoration, VLMs can reason over degraded text using both visual and linguistic priors, enabling them to provide explicit textual guidance for the restoration process. This makes them a promising alternative to prior complex multi-module text-guided pipelines. However, when the degraded text is semantically unrelated to the scene context or when visual cues are insufficient to exploit VLM priors, the VLM’s text predictions may become unreliable. To address this limitation, we incorporate a Text Spotting Module (TSM) into our restoration framework~\cite{min2025text} and train it on diffusion features to generate OCR predictions at each denoising timestep. This design enables the VLM to leverage these intermediate OCR predictions to iteratively refine its initial guidance, correcting potential errors and improving the accuracy of the textual information supplied to the restoration model. The combined use of the VLM and TSM provides robust and faithful textual guidance, even under severe degradation conditions.

\begin{figure}[t]
    \centering
    \includegraphics[width=\linewidth]{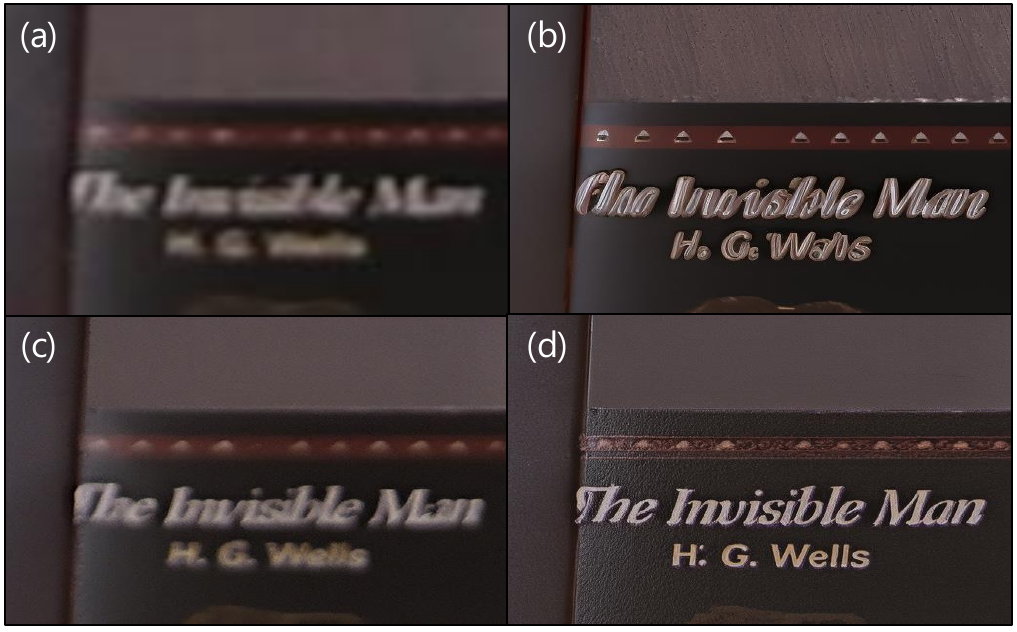}
    \vspace{-20pt}
    \caption{\textbf{Text restoration with ground-truth (GT) guidance.} 
    (a) LQ input, (b) TeReDiff~\cite{min2025text}, (c) DiT4SR~\cite{duan2025dit4sr}, and (d) UniT (Ours). 
    Despite being provided with the GT text prompt \textit{“The Invisible Man H. G. Wells”}, the UNet-based model (b) fails to correctly reconstruct the textual content. In contrast, the DiT-based models (c) and (d) more effectively utilize the explicit text guidance, with our method (d) producing the most accurate text restoration.}
    \label{fig:restoration_gtprompt}
    \vspace{-5pt}
\end{figure}

Lastly, even with accurate textual guidance, diffusion-based architectures may not fully exploit these cues to perform text restoration. Our experiments indicate that prior UNet-based models possess limited capacity for precise text restoration, whereas state-of-the-art diffusion transformers (DiTs)~\cite{duan2025dit4sr, peebles2023scalable, esser2024scaling, flux2024, chen2023pixart} achieve high-fidelity recovery of intricate, high-resolution text structures through their full attention mechanisms. These characteristics render DiTs particularly well-suited to handle complex text regions, highlighting their potential to deliver more accurate and consistent text image restoration compared to earlier UNet-based approaches (Fig.~\ref{fig:restoration_gtprompt}).

To this end, we present \textbf{UniT} (\textbf{Uni}fied Diffusion-\textbf{T}ransformer), a unified text-aware image restoration framework that combines a VLM~\cite{liu2023visual, Qwen2.5-VL}, TSM~\cite{zhang2022text}, and DiT~\cite{esser2024scaling, duan2025dit4sr} in an iterative fashion for high-fidelity, text-aware image restoration (Fig.~\ref{fig:unit_teaser}). In UniT, each component serves a complementary purpose: the VLM extracts textual content from degraded images to provide initial text guidance, the TSM generates intermediate OCR predictions at each denoising step to enable the VLM to iteratively correct potential errors, and the DiT backbone leverages its representational power to recover fine-grained text details while suppressing text hallucinations. Experiments on the SA-Text and Real-Text benchmarks~\cite{min2025text} demonstrate that UniT faithfully reconstructs degraded text, substantially reduces text hallucinations, and achieves state-of-the-art end-to-end (E2E) F1-score performance in text-aware image restoration.

\section{Related Work}
\label{sec:2_related}

\paragraph{Diffusion models for image restoration.}
Recent advances in high-quality image synthesis using diffusion models~\cite{ramesh2021zero, rombach2022high, podell2023sdxl} have motivated their adoption in image restoration~\cite{lin2024diffbir, mei2024codi, sahak2023denoising, saharia2022image, chen2024faithdiff, ai2024dreamclear, duan2025dit4sr, wu2024seesr}. Compared to GAN-based approaches~\cite{wang2018esrgan, wang2021real, zhang2021blind, che2016mode}, which often suffer from unstable training and mode collapse, diffusion models provide stable optimization, robustness, and strong generalization through their iterative denoising mechanism. Recently, DiTs~\cite{peebles2023scalable, esser2024scaling} have emerged as effective alternatives to UNet backbones, with their attention mechanism enabling global receptive fields, scalable computation, and stronger feature representations, yielding superior performance in vision tasks~\cite{peebles2023scalable, esser2024scaling, chen2023pixart, chen2024gentron, flux2024, ai2024dreamclear, duan2025dit4sr}. This work also builds upon a DiT-based diffusion model~\cite{duan2025dit4sr} to leverage its rich generative priors and strong representations, often guided by explicit textual cues, for effective restoration.
\vspace{-5pt}

\begin{table}
\centering
\scriptsize 
\setlength{\tabcolsep}{3pt} 
    \begin{tabular}{l|c|c|c|c}
        \toprule
            Model & Incorrect (\%) & Partial (\%) & Exact (\%) & Partial+Exact (\%)  \\    
        \midrule
            LLaVA (7B) & 65.70 & \underline{12.10}  & 22.20 &  34.30 \\
            LLaVA (13B) & 67.10 & 11.70 & 21.20 & 32.90 \\
        \midrule
            Qwen2.5-VL (3B) & 65.50 & 10.20 & 24.30 & 34.50 \\
            Qwen2.5-VL (7B) & \textbf{60.80} & \textbf{12.40} & \underline{26.80} & \textbf{39.20} \\
            Qwen2.5-VL (32B) & 62.90 & 11.40 & 25.70 & 37.10 \\
            Qwen2.5-VL (72B) & \underline{61.80} & 11.38 & \textbf{26.82} & \underline{38.20} \\
        \bottomrule
    \end{tabular}
    \vspace{-7pt}
    \caption{\textbf{Text extraction performance of VLMs from LQ image.} We evaluate LLaVA1.5~\cite{liu2023visual} and Qwen2.5-VL~\cite{Qwen2.5-VL} of varying sizes on SA-Text (lv3)~\cite{min2025text}, which represents the most challenging degradation setting where texts are hardly visible. Outputs are classified as Incorrect, Partially correct, or Exactly correct. Notably, Qwen2.5-VL 7B achieves the best performance despite its smaller size.}
    \vspace{-10pt}
    \label{tab:vlm_lq_extract}
\end{table}
\paragraph{Text-aware image restoration.}
Text-aware image restoration (TAIR) aims to recover all degraded textual content contained in full scene low-quality images. A related subtask, Scene Text Image Super-Resolution (STISR), also restores degraded text but operates on word-level cropped images which contain only one line of texts. Early STISR methods used CNN-~\cite{dong2015boosting} and GAN-based approaches with text-aware losses~\cite{wang2019textsr, xu2017learning, ledig2017photo}, followed by residual-block architectures~\cite{mou2020plugnet}, high-frequency enhancement techniques~\cite{wang2020scene}, and transformer-based methods leveraging text priors~\cite{ma2023text, ma2022text}, multimodal cues~\cite{zhao2022c3}, or position- and content-aware losses~\cite{chen2022text}. Diffusion-based STISR frameworks have also been proposed~\cite{liu2023textdiff, noguchi2024scene, singh2024dcdm, ye2025textsr}. In contrast, TAIR focuses on full-scene images, where text exhibits variations in scale, orientation, and font, necessitating joint reasoning over local character patterns and global scene context. Prior methods have addressed this challenge by incorporating segmentation maps~\cite{hu2025text, luo2025restore}, multi-task learning frameworks~\cite{hu2025text, min2025text}, and text-aware learning via a text spotting module~\cite{min2025text}. Given that this setting more accurately reflects real-world scenarios, we adopt the TAIR formulation.
\vspace{-5pt}

\paragraph{VLMs for image restoration.}
Image restoration~\cite{dabov2007color, cai2016dehazenet, chen2021all, ren2019progressive, tao2018scale, guo2016lime} is an inherently ill-posed problem, where a single degraded observation may admit numerous plausible underlying clean images due to the loss, corruption, or ambiguity of essential visual cues. This lack of sufficient constraints prevents the formulation of a unique and well-defined solution, making the recovery of the original content fundamentally ambiguous. To address this challenge, prior diffusion-based restoration models~\cite{wu2024seesr, chen2024faithdiff, ai2024dreamclear, yu2024scaling, duan2025dit4sr} employ a captioner to provide general scene context as textual guidance during the diffusion process. For example, SeeSR~\cite{wu2024seesr} trains a degradation-aware prompt extractor to obtain robust scene-context information, while FaithDiff~\cite{chen2024faithdiff}, SUPIR~\cite{yu2024scaling}, DreamClear~\cite{ai2024dreamclear}, and DiT4SR~\cite{duan2025dit4sr} utilize an external visual language model~\cite{liu2023visual} to supply semantic guidance, thereby improving restoration performance. Despite not being trained on degraded images, visual language models possess rich visual and linguistic priors that enable them to generate reliable scene-context captions for low-quality inputs. Motivated by these observations, our approach leverages the rich priors of VLMs~\cite{liu2023visual, Qwen2.5-VL} to extract degraded textual content and provide explicit textual guidance for effective text-aware image restoration.

\begin{table}[t]
\centering
\footnotesize
\resizebox{\columnwidth}{!}{
    \begin{tabular}{c|c|c|c}
    \toprule
    Dataset & Model & E2E(None) F1 ($\uparrow$) & E2E(Full) F1 ($\uparrow$) \\
    \midrule
    \multirow{2}{*}{SA-Text (lv1)} & TeReDiff~\cite{min2025text} & 34.71 & 44.87 \\
                              & DiT4SR~\cite{duan2025dit4sr} & \textbf{43.20} & \textbf{54.45} \\
    \midrule
    \multirow{2}{*}{SA-Text (lv2)} & TeReDiff~\cite{min2025text} & 33.31 & 43.4 \\
                               & DiT4SR~\cite{duan2025dit4sr} & \textbf{39.33} & \textbf{51.36} \\
    \midrule
    \multirow{2}{*}{SA-Text (lv3)} & TeReDiff~\cite{min2025text} & 27.91 & 37.76 \\
                               & DiT4SR~\cite{duan2025dit4sr} & \textbf{30.87} & \textbf{41.88} \\
    \midrule
    \multirow{2}{*}{RealText} & TeReDiff~\cite{min2025text} & 54.06 & 60.74 \\
                               & DiT4SR~\cite{duan2025dit4sr} & \textbf{58.94} & \textbf{66.35} \\
    \bottomrule
    \end{tabular}
}
\vspace{-7pt}
\caption{\textbf{Text restoration with ground-truth (GT) guidance.} 
We compare the ability of two representative diffusion backbones, the UNet-based TeReDiff~\cite{min2025text} and the DiT-based DiT4SR~\cite{duan2025dit4sr}, to leverage explicit GT textual prompts for text restoration. GT text is provided as guidance, and the restored outputs are assessed using a text spotting model to quantify how effectively each architecture exploits textual cues for high-fidelity text restoration.}
\label{tab:dit_unet_gtprompt}
\end{table}

\section{Motivation and Overview}

Restoring textual content from LQ images is a challenging task due to severe degradation and weak visual cues. We find that VLMs, equipped with rich visual-linguistic priors, can effectively extract degraded text from LQ images and provide explicit textual guidance for text restoration (Sec.~\ref{method_subsec:vlm_text_extract}). 

However, their predictions may become unreliable when the degraded text is semantically unrelated to the scene or when visual and linguistic cues are insufficient. To address this limitation, we train a text spotting module (TSM) within the diffusion restoration framework to generate intermediate OCR outputs at each denoising timestep, enabling VLM self-correction and thereby providing more accurate textual guidance. The complementary use of our VLM and TSM demonstrates that OCR outputs from the TSM can effectively facilitate VLM self-correction, yielding more faithful textual guidance (Sec.~\ref{method_subsec:learn_tsm}). 

We further examine the capacity of diffusion backbone architectures to fully exploit textual cues for high-fidelity text reconstruction by providing ground-truth (GT) textual guidance to representative UNet~\cite{min2025text} and DiT~\cite{duan2025dit4sr} based restoration models. Our results indicate that DiT-based architectures achieve superior text restoration compared to UNet-based models, highlighting the critical role of backbone selection in leveraging textual guidance (Sec.~\ref{method_subsec:dit_backbone}).

\subsection{VLMs for Text Extraction}
\label{method_subsec:vlm_text_extract}
\begin{figure}[t]
    \centering
    \includegraphics[width=\linewidth]{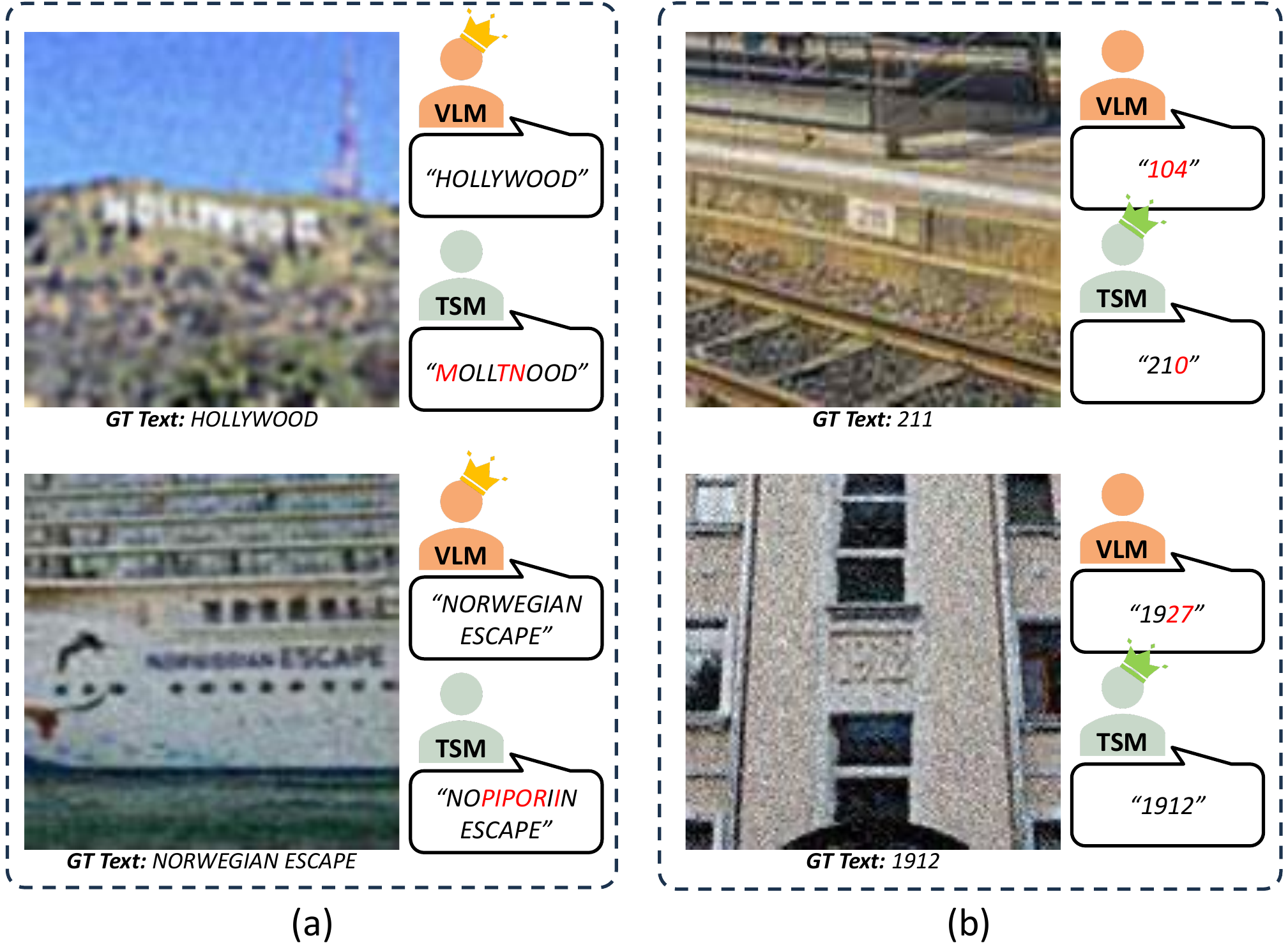}
    \vspace{-20pt}
    \caption{\textbf{Distinct strengths of VLM and TSM.}
    (a) The VLM~\cite{Qwen2.5-VL} demonstrates superior performance on heavily degraded complex words (e.g., HOLLYWOOD, NORWEGIAN ESCAPE) by exploiting visual-linguistic priors to infer plausible full-word predictions that the TSM fails to recover.
    (b) The TSM achieves higher accuracy on simpler, character-level text (e.g., 211, 1912), producing stable symbol-level predictions when local glyph structures remain partially visible.}
    \label{fig:advantage_vlm_tsm}
    \vspace{-5pt}
\end{figure}

We employ VLMs to extract textual information from LQ images, enabling them to serve as explicit textual guidance for the restoration model. To identify the most suitable VLM for this task, we evaluate two representative architectures, LLaVA1.5~\cite{liu2023visual} and Qwen2.5-VL~\cite{Qwen2.5-VL}, across multiple model scales, comparing their ability to recover degraded text on the SA-Text (lv3) dataset~\cite{min2025text}, which comprises 1,000 images subjected to severe level-3 degradation. Both models are queried using the instruction ``OCR this image and transcribe only the English text,'' and their outputs are evaluated against ground-truth annotations, with results categorized as Incorrect, Partially Correct, or Exactly Correct. The findings, summarized in Tab.~\ref{tab:vlm_lq_extract}, indicate that increased model size does not necessarily correlate with improved extraction performance. Notably, Qwen2.5-VL consistently outperforms LLaVA, with its 7B variant achieving higher text extraction accuracy than even the 72B model. Fig.~\ref{fig:advantage_vlm_tsm} presents several examples illustrating the LQ text extraction capabilities of Qwen2.5-VL 7B, demonstrating its ability to provide robust textual guidance that facilitates effective text restoration. Based on these empirical results, we adopt Qwen2.5-VL 7B as the VLM component in our framework. Additional details regarding the classification methodology, LQ extraction results for different query instructions, and visualizations of level-3 degradations in SA-Text~\cite{min2025text} are provided in the supplementary material.

\subsection{Learning a Text Spotting Module}
\label{method_subsec:learn_tsm}
\begin{figure}[t]
    \centering
    \includegraphics[width=\linewidth]{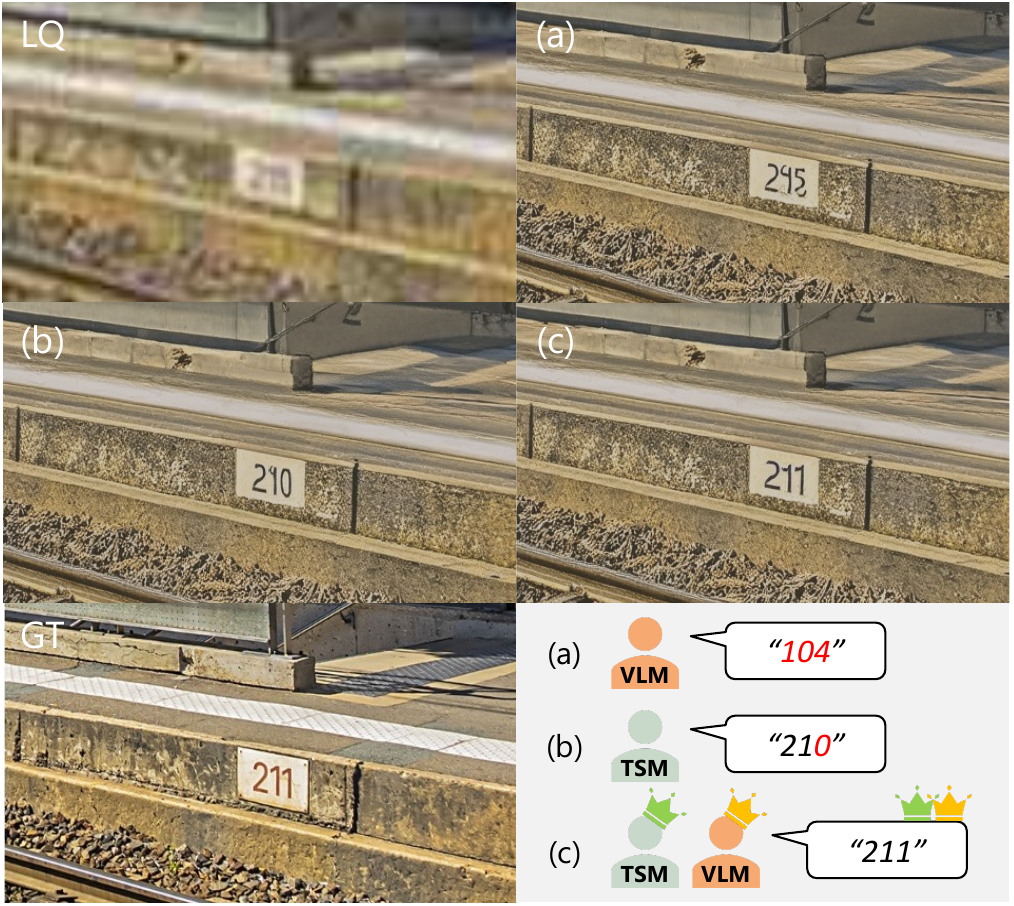}
    \vspace{-20pt}
    \caption{\textbf{Complementary usage of VLM and TSM for effective text restoration.} Text restoration using (a) only the VLM (text prediction: 104) fails because it misinterprets simpler, partially visible characters despite its strength in inferring complex words. (b) The TSM (text prediction: 210) yields partial improvement but still shows errors. (c) Joint usage of both TSM and VLM enables VLM self-correction (text prediction: 211), resulting in faithful textual guidance for accurate text restoration.}
    \label{fig:complementary_vlm_tsm}
    \vspace{-5pt}
\end{figure}

While VLMs provide rich visual-linguistic priors and can effectively extract textual content from LQ images, their predictions may become unreliable in scenarios where the degraded text is semantically unrelated to the scene context or where visual cues are insufficient to leverage their priors. To address this limitation, we enable VLMs to self-correct initial incorrect predictions using intermediate denoising steps of the diffusion model. Specifically, we train a TSM within the diffusion-based restoration framework to generate intermediate OCR predictions at each denoising timestep. These intermediate predictions allow the VLM to iteratively refine its text guidance in conjunction with the LQ image, progressively improving the accuracy of textual cues provided to the restoration model throughout the denoising process. We observe that this is possible as the VLM and TSM have distinct strengths for LQ text extraction: the VLM can leverage its rich priors to predict complex degraded texts, while the TSM can accurately extract highly localized simple characters and numerical text (see Fig.~\ref{fig:advantage_vlm_tsm}). By jointly using both VLM and TSM, we exploit their complementary strengths, leveraging intermediate OCR predictions from the TSM to stimulate VLM self-correction and provide faithful textual guidance, thereby achieving accurate text restoration, as shown in Fig.~\ref{fig:complementary_vlm_tsm} and Tab.~\ref{tab:abl_vlm_tsm}.

\begin{figure*}[t!]
    \centering
    \includegraphics[width=\linewidth]{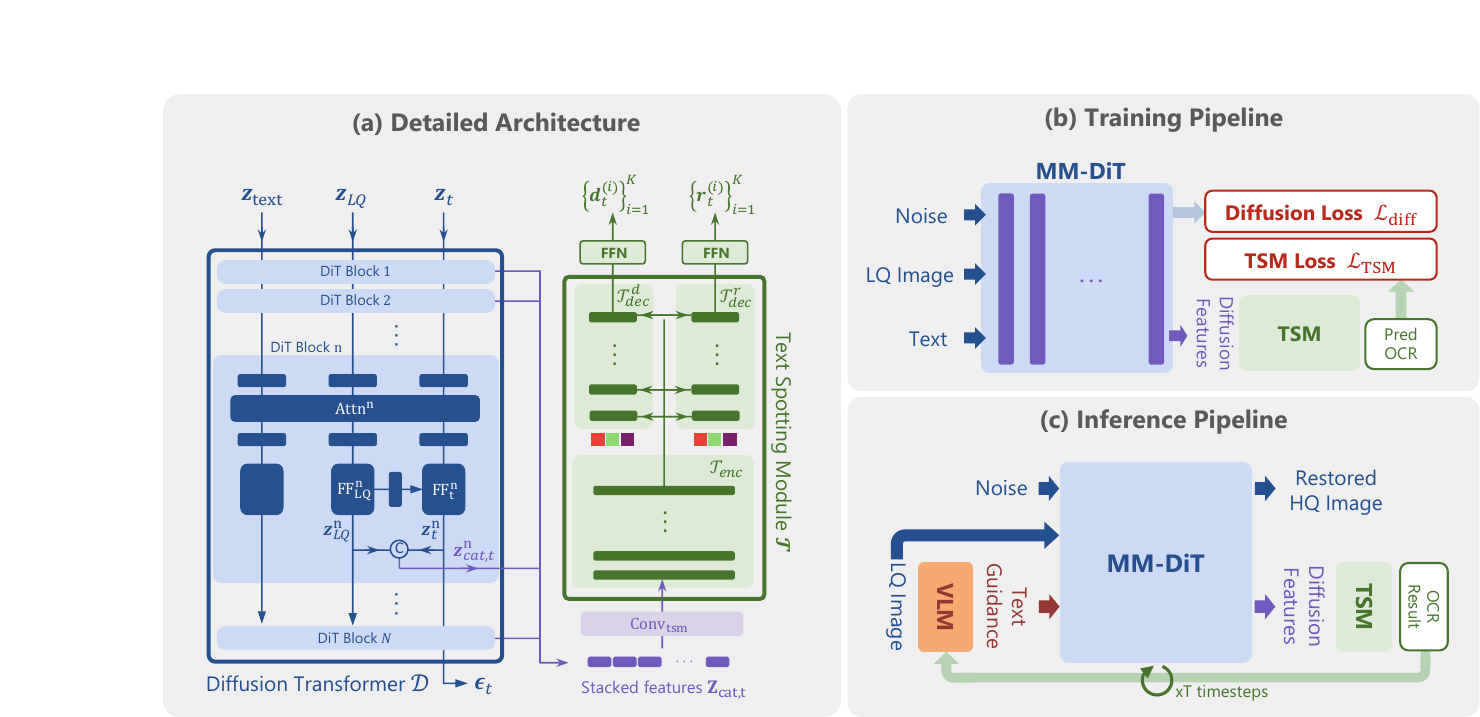}
    \vspace{-20pt}
    \caption{\textbf{UniT framework overview.} (a) Detailed architecture of the DiT-based image restoration model integrated with the TSM~\cite{zhang2022text}. (b) The TSM is trained on the diffusion features extracted from the DiT backbone. (c) During inference, the VLM~\cite{Qwen2.5-VL} utilizes the intermediate OCR predictions from the TSM to refine its initial text guidance, thereby improving text restoration fidelity.}
    \vspace{-10pt}
    \label{fig:framework}
\end{figure*}
\subsection{DiT Backbone for Text Restoration}
\label{method_subsec:dit_backbone}
Even with accurate textual cues, diffusion-based architectures may not fully leverage or comprehend these signals to enhance text restoration. Therefore, we investigate the capacity of different diffusion backbone architectures to effectively exploit textual guidance for text-aware image restoration. We evaluate representative models from both UNet- and DiT-based diffusion restoration frameworks. For the UNet-based representative, we employ TeReDiff~\cite{min2025text}, which outperforms previous UNet-based diffusion models on text-aware image restoration. For the DiT-based diffusion backbone, we adopt DiT4SR~\cite{duan2025dit4sr}, currently the best-performing DiT-based model for image restoration, at the time of writing. We assess the two architectures’ ability to utilize textual cues by providing ground-truth texts of LQ images to both models for reconstructing degraded textual content. As shown in Fig.~\ref{fig:restoration_gtprompt} and Tab.~\ref{tab:dit_unet_gtprompt}, although DiT4SR is not trained on additional text-centric restoration datasets, its ability to leverage explicit textual guidance surpasses that of the prior UNet based state-of-the-art model TeReDiff on the SA-Text and Real-Text benchmarks~\cite{min2025text}, highlighting the importance of the diffusion backbone in effectively utilizing textual cues for faithful text restoration.

\section{Unified Diffusion Transformer Framework}

Fig.~\ref{fig:framework} presents an overview of the proposed UniT framework, which comprises three primary components: a VLM, a TSM, and a DiT backbone for high-fidelity text restoration. Each module serves a distinct yet complementary role. The VLM extracts textual content from degraded images to provide initial guidance, the TSM generates intermediate OCR predictions at each denoising step to enable iterative VLM self-correction, and the DiT backbone leverages its strong representational power to recover fine-grained text details while suppressing hallucinations. 

We begin by presenting the formulation of the diffusion-based denoising objective for text restoration in Subsection~\ref{sec_unit:prelim}, then detail the training objective in Subsection~\ref{sec_unit:training_pipeline}, and finally describe the inference pipeline in Subsection~\ref{sec_unit:inference_pipeline}.

\subsection{Task Formulation}
\label{sec_unit:prelim}

Given a low-quality (LQ) image $\mathbf{x}_{LQ}$, our objective is to restore a high-quality (HQ) image $\mathbf{x}_{HQ}$ based on the following the probability:
\begin{equation}
    p(\mathbf{x}_{HQ} \mid \mathbf{x}_{LQ}).
\end{equation}
To implement this with diffusion models, starting from gaussian noise $\mathbf{z}_T$, the denoising sequence $(\mathbf{z}_T, \mathbf{z}_{T-1}, \dots, \mathbf{z}_0)$ is modeled as $p(\mathbf{z}_T, \mathbf{z}_{T-1}, \dots, \mathbf{z}_0) = p(\mathbf{z}_T) \prod_{t=1}^{T} p(\mathbf{z}_{t-1} \mid \mathbf{z}_t)$ under the Markovian assumption. However, as $p(\mathbf{z}_{t-1} \mid \mathbf{z}_t)$ is generally intractable, it is approximated by a diffusion model parameterized by $\theta$ as $p_\theta(\mathbf{z}_{t-1} \mid \mathbf{z}_t) \approx p(\mathbf{z}_{t-1} \mid \mathbf{z}_t)$. For image restoration, the model is further conditioned on the LQ latent $\mathbf{z}_{LQ}$, yielding
\begin{equation}
    p_\theta(\mathbf{z}_{t-1} \mid \mathbf{z}_t, \mathbf{z}_{LQ}).
\end{equation}

\paragraph{Textual guidance.}
To further constrain the solution space and mitigate text hallucination for high-fidelity text-aware image restoration, text captions can be extracted from the LQ image using a vision-language model (VLM) parameterized by $\phi$ and a text spotting module (TSM) parameterized by $\psi$ as $p_\phi(\mathbf{c}_{\mathrm{vlm}} \mid \mathbf{x}_{LQ})$ and $p_\psi(\mathbf{c}_{\mathrm{tsm}} \mid \mathbf{x}_{LQ})$, respectively. The extracted captions $\mathbf{c}_{\mathrm{vlm}}$ and $\mathbf{c}_{\mathrm{tsm}}$ are subsequently encoded into $\mathbf{z}_{\mathrm{text}}$, which is incorporated as textual guidance in the denoising process:
\begin{equation}
    p_\theta(\mathbf{z}_{t-1} \mid \mathbf{z}_t, \mathbf{z}_{LQ}, \mathbf{z}_{\mathrm{text}}).
\end{equation}

\paragraph{Final denoising objective.}
Incorporating both the LQ latent $\mathbf{z}_{LQ}$ and the text-guidance latent $\mathbf{z}_{\mathrm{text}}$, the joint distribution of the denoising timesteps is expressed as:
\begin{equation}
    p(\mathbf{z}_T, \mathbf{z}_{T-1}, \dots, \mathbf{z}_0)
    = p(\mathbf{z}_T)\, \prod_{t=1}^{T} 
    p_\theta(\mathbf{z}_{t-1} \mid \mathbf{z}_t, \mathbf{z}_{LQ}, \mathbf{z}_{\mathrm{text}}),
\end{equation}
where $\mathbf{z}_T \sim \mathcal{N}(\mathbf{0}, \mathbf{I})$ and $\mathbf{z}_{LQ} = \mathrm{VAE_{Enc}}(\mathbf{x}_{LQ})$, where $\mathrm{VAE_{Enc}}$ denotes a VAE encoder~\cite{kingma2013auto}. Refer to the supplementary material for further details.

\begin{table*}[!t]
    \centering
    \resizebox{\textwidth}{!}{
        \begin{tabular}{ c|l|ccccc|ccccc }
            \toprule
            \multirow{3}{*}{\textbf{Deg. Level}} &
            \multirow{3}{*}{\textbf{Model}} &
            \multicolumn{5}{c|}{\textbf{ABCNet v2~\citep{liu2021abcnet}}} &
            \multicolumn{5}{c}{\textbf{TESTR~\citep{zhang2022text}}} \\ 
            
            \cmidrule(lr){3-7}
            \cmidrule(lr){8-12}
            & & 
            \multicolumn{3}{c}{Detection} &
            \multicolumn{2}{c|}{End-to-End} &
            \multicolumn{3}{c}{Detection} &
            \multicolumn{2}{c}{End-to-End} \\
            
            \cmidrule(lr){3-5}
            \cmidrule(lr){6-7}
            \cmidrule(lr){8-10}
            \cmidrule(lr){11-12}
            & & Precision($\uparrow$) & Recall($\uparrow$) & F1-Score($\uparrow$) & None($\uparrow$) & Full($\uparrow$) & Precision($\uparrow$) & Recall($\uparrow$) & F1-Score($\uparrow$) & None($\uparrow$) & Full($\uparrow$) \\ 
            \midrule
            - & HQ (GT) & 92.16 & 86.85 & 89.43 & 71.79 & 82.05 & 92.59 & 87.81 & 90.13 & 75.90 & 84.18 \\ 
            \midrule
            
            \multirow{13}{*}{Level1} 
            & LQ (Lv1) & 89.79 & 29.51 & 44.42 & 24.29 & 34.25 & 84.01 & 30.24 & 44.47 & 25.93 & 34.73 \\ 
            \cmidrule{2-12}
            & Real-ESRGAN~\citep{wang2021real} & 83.79 & 43.34 & 57.13 & 21.45 & 30.12 & 85.19 & 41.98 & 56.24 & 22.90 & 31.52 \\
            & SwinIR~\citep{liang2021swinir} & {84.95} & 40.93 & 55.25 & 22.70 & 31.26 & \underline{87.93} & 39.62 & 54.63 & 25.50 & 33.75 \\ 
            & ResShift~\citep{yue2023resshift} & 81.93 & 40.07 & 53.82 & 20.40 & 28.74 & \textbf{88.47} & 35.81 & 50.98 & 22.39 & 30.14 \\
            & StableSR~\citep{wang2024exploiting} & 77.90 & 55.44 & 64.78 & 21.93 & 29.29 & 84.44 & {50.68} & 63.34 & 24.02 & 31.84 \\ 
            & DiffBIR~\citep{lin2024diffbir} & 76.29 & 56.44 & 64.88 & {23.14} & {32.73} & 84.00 & 52.13 & 64.34 & {25.51} & {35.47} \\
            & SeeSR~\citep{wu2024seesr} & 70.00 & \textbf{61.88} & {65.69} & 20.16 & 28.63 & 78.85 & \underline{55.76} & {65.32} & 23.31 & 32.82 \\
            & SUPIR~\citep{yu2024scaling} & 43.64 & 49.46 & 46.37 & 14.58 & 19.34 & 53.02 & 46.19 & 49.37 & 17.44 & 22.29 \\

            & FaithDiff~\citep{chen2024faithdiff} & 69.16 & \underline{61.51} & 65.12 & 20.44 & 27.78 & 78.80 & \textbf{57.12} & \underline{66.23} & 22.50 & 31.59 \\ 
            
            & TeReDiff~\cite{min2025text} & \underline{85.29} & {58.34} & \textbf{69.29} & \underline{26.59} & \underline{35.69} & 87.50 & \underline{54.90} & \textbf{67.47} & {28.19} & \underline{36.99} \\

            
            & DiT4SR~\cite{duan2025dit4sr} & 73.68	& 58.11	& 64.98 & 26.56 & 34.31 & 81.47	& 54.81	& 65.53 & \underline{28.94} & 36.48 \\ 

            \cmidrule{2-12} 
            
            & \textbf{UniT (Ours)} & \textbf{85.89}	& 55.49	& \underline{67.42} & \textbf{30.96} & \textbf{39.93} & 87.47	& 51.59	& 64.90 & \textbf{31.99} & \textbf{41.52} \\

            \midrule
            \midrule

            \multirow{13}{*}{Level2} 
            & LQ (Lv2) & 87.67 & 22.89 & 36.30 & 20.49 & 27.82 & 78.45 & 23.93 & 36.68 & 20.49 & 27.37 \\ 
            \cmidrule{2-12}
            & Real-ESRGAN~\citep{wang2021real} & {81.42} & 41.12 & 54.64 & 18.31 & 24.88 & 84.92 & 38.80 & 53.27 & 19.29 & 27.50 \\ 
            & SwinIR~\citep{liang2021swinir} & 80.14 & 37.31 & 50.91 & 17.82 & 24.93 & {85.43} & 34.81 & 49.47 & 19.07 & 26.99 \\ 
            & ResShift~\citep{yue2023resshift} & 81.11 & 35.22 & 49.12 & 17.89 & 25.54 & 85.18 & 32.05 & 46.57 & 17.26 & 26.09 \\ 
            & StableSR~\citep{wang2024exploiting} & 75.49 & 51.95 & 61.55 & 19.55 & 26.69 & 79.03 & 48.87 & 60.39 & 20.06 & 27.68 \\ 
            & DiffBIR~\citep{lin2024diffbir} & 72.96 & 53.94 & 62.03 & {19.60} & {27.52} & 79.69 & 50.50 & 61.82 & 21.64 & {30.36} \\ 
            & SeeSR~\citep{wu2024seesr} & 68.93 & \textbf{60.65} & \underline{64.53} & 19.48 & 26.72 & 77.50 & \textbf{54.49} & \underline{63.99} & {21.83} & 29.17 \\ 
            & SUPIR~\citep{yu2024scaling} & 42.01 & 45.65 & 43.75 & 13.21 & 17.42 & 53.54 & 43.25 & 47.84 & 15.50 & 19.96 \\ 

            & FaithDiff~\citep{chen2024faithdiff} & 66.62 & \underline{59.34} & 62.77 & 18.94 & 25.99 & 76.16 & \underline{54.17} & 63.31 & 20.98 & 28.56 \\ 
            
            & TeReDiff~\cite{min2025text} & \underline{83.02} & {56.30} & \textbf{67.10} & \underline{24.42} & \underline{33.23} & \underline{86.95} & {52.86} & \textbf{65.75} & \underline{26.39} & \underline{35.13} \\

            
            & DiT4SR~\cite{duan2025dit4sr} & 69.22 & 54.03 & 60.69 & 23.27 & 29.99 & 76.56 & 51.09 & 61.28 & 25.39 & 32.63 \\ 

            \cmidrule{2-12} 
            
            & \textbf{UniT (Ours)} & \textbf{83.73} & 50.86 & 63.28 & \textbf{28.43} & \textbf{36.83}  & \textbf{87.53} & 48.69	& 62.57 & \textbf{30.12} & \textbf{37.98} \\ 
            
            \midrule
            \midrule

            \multirow{13}{*}{Level3} 
            & LQ (Lv3) & 85.38 & 13.24 & 22.92 & 12.17 & 16.95 & 76.01 & 15.37 & 25.57 & 12.52 & 17.72 \\ 
            \cmidrule{2-12}
            & Real-ESRGAN~\citep{wang2021real} & 72.48 & 28.65 & 41.07 & 11.89 & 16.37 & 76.51 & 27.02 & 39.93 & 12.13 & 18.22 \\
            & SwinIR~\citep{liang2021swinir} & 74.41 & 25.57 & 38.06 & 11.27 & 16.33 & {78.38} & 24.16 & 36.94 & 11.85 & 17.12 \\ 
            & ResShift~\citep{yue2023resshift} & 75.00 & 22.57 & 34.70 & 10.80 & 15.19 & 81.10 & 20.04 & 32.13 & 9.89 & 15.63 \\
            & StableSR~\citep{wang2024exploiting} & 67.63 & 38.08 & 48.72 & 13.34 & 18.91 & 72.21 & 35.22 & 47.35 & 13.65 & 19.68 \\ 
            & DiffBIR~\citep{lin2024diffbir} & 59.30 & 42.20 & 49.31 & {13.88} & {19.39} & 72.27 & 38.98 & 50.65 & 15.61 & {22.67} \\
            & SeeSR~\citep{wu2024seesr} & 55.06 & \underline{46.83} & {50.61} & 13.38 & 18.47 & 64.95 & \underline{43.93} & \underline{52.41} & {14.93} & 20.88 \\ 
            & SUPIR~\citep{yu2024scaling} & 31.05 & 34.72 & 32.78 & 9.07 & 11.77 & 40.78 & 32.77 & 36.34 & 11.21 & 14.02 \\

            & FaithDiff~\citep{chen2024faithdiff} & 56.04 & \textbf{47.91} & 51.66 & 13.69 & 19.01 & 69.44 & \textbf{45.01} & 54.62 & 15.40 & 21.18 \\

            & TeReDiff~\cite{min2025text} & \underline{81.76} & {44.11} & \textbf{57.30} & \underline{19.61} & \underline{27.50} & \underline{84.50} & {42.02} & \textbf{56.13} & \underline{19.92} & \underline{28.34} \\ 

            
            & DiT4SR~\cite{duan2025dit4sr} & 60.67 & 43.29 & 50.53 & 16.61 & 22.28 & 70.57 & 40.44 & 51.41 & 17.98 & 24.55 \\ 

            \cmidrule{2-12} 
            
            & \textbf{UniT (Ours)} & \textbf{83.14}	& 38.21	& \underline{52.36} & \textbf{21.12} & \textbf{28.63} & \textbf{86.43}	& 36.94	& 51.76 & \textbf{22.86} & \textbf{29.53} \\ 

            \bottomrule
        \end{tabular}
    }
    \vspace{-7pt}
    \caption{\textbf{Text restoration quantitative results on three degradation levels of SA-Text~\cite{min2025text}.} Text restoration performance is evaluated through text detection and recognition by two text spotting modules~\cite{liu2021abcnet, zhang2022text}. Higher levels indicate stronger degradation settings. ‘None’ and ‘Full’ denote recognition without and with a full lexicon, respectively. Best and second-best results are shown in \textbf{bold} and \underline{underlined}.}
    \vspace{-5pt}
    \label{tab:metric_ts_satext}
\end{table*}

\subsection{Architecture Overview}
\label{sec_unit:training_pipeline}
Given an LQ image $\mathbf{x}_{LQ} \in \mathbb{R}^{H \times W \times 3}$, our objective is to restore its HQ counterpart $\mathbf{x}_{HQ} \in \mathbb{R}^{H \times W \times 3}$. Using a VLM~\cite{Qwen2.5-VL}, we first extract the degraded textual content $\mathbf{c}_{\mathrm{vlm}}$ from $\mathbf{x}_{LQ}$ and encode it as $\mathbf{z}_{\mathrm{text}}$. Following DiT4SR~\cite{duan2025dit4sr}, each MM-DiT block includes an additional LQ branch, where, at denoising timestep $t$, the denoiser jointly operates on the noisy latent $\mathbf{z}_t \in \mathbb{R}^{L \times D}$, the LQ latent $\mathbf{z}_{LQ} \in \mathbb{R}^{L \times D}$, and the text guidance latent $\mathbf{z}_{\mathrm{text}} \in \mathbb{R}^{M \times D}$. These three latents are propagated through $N$ DiT blocks, each comprising a self-attention layer $\mathrm{Attn}_{n}$ and a feed-forward layer $\mathrm{FF}_{n}$ for $n \in \{1, \dots, N\}$.

Furthermore, rather than relying on a separate feature extractor, the TSM is trained by directly leveraging the internal representations of the diffusion model. Specifically, internal features from all MM-DiT blocks are extracted and concatenated into $\mathrm{\textbf{Z}}_{\mathrm{cat},t} = (\mathbf{z}_{\mathrm{cat},t}^1, \dots, \mathbf{z}_{\mathrm{cat},t}^{N})$,
where each $\mathbf{z}_{\mathrm{cat},t}^i \in \mathbb{R}^{2L \times D}$ is formed by combining the feed-forward outputs of $\mathbf{z}_{LQ}^{i}$ and $\mathbf{z}_t^{i}$. Before entering the TSM, $\mathrm{\textbf{Z}}_{\mathrm{cat},t}$ is processed by a learnable convolution layer $\mathrm{Conv}_{\mathrm{tsm}}(\cdot)$. The TSM includes an encoder $\mathcal{T}_{\mathrm{enc}}$ and two decoders, $\mathcal{T}_{\mathrm{dec}}^\mathrm{d}$ and $\mathcal{T}_{\mathrm{dec}}^\mathrm{r}$, which produce $K$ text detection outputs $\{\textbf{d}_t^{(i)}\}_{i=1}^K$ and $K$ recognition outputs $\{\textbf{r}_t^{(i)}\}_{i=1}^K$. Each detection $\textbf{d}^{(i)}$ is a polygon defined by control points, and each recognition $\textbf{r}^{(i)}$ is a sequence of predicted characters.

\subsection{Training}

During training, the restoration module parameterized by $\theta$ is optimized using the standard diffusion loss from SD3~\cite{esser2024scaling} and DiT4SR~\cite{duan2025dit4sr}, while the TSM parameterized by $\psi$ is trained with the conventional text spotting loss, comprising detection and recognition components. VLM parameterized by $\phi$ is frozen to preserve its generalization ability. Detailed definitions of all loss functions are provided in the supplementary material.

\begin{table*}[t]
    \centering
    \resizebox{\linewidth}{!}{
        \begin{tabular}{ l|ccccc|ccccc } 
            \toprule
            \multirow{3}{*}{\textbf{Model}} &
            \multicolumn{5}{c|}{\textbf{ABCNet v2~\citep{liu2021abcnet}}} & 
            \multicolumn{5}{c}{\textbf{TESTR~\citep{zhang2022text}}} \\ 
    
            \cmidrule(lr){2-6}
            \cmidrule(lr){7-11}
            &
            \multicolumn{3}{c}{Detection} &
            \multicolumn{2}{c|}{End-to-End} &
            \multicolumn{3}{c}{Detection} &
            \multicolumn{2}{c}{End-to-End} \\

            \cmidrule(lr){2-4}
            \cmidrule(lr){5-6}
            \cmidrule(lr){7-9}
            \cmidrule(lr){10-11}
          
            & Precision($\uparrow$) & Recall($\uparrow$) & F1-Score($\uparrow$) & None($\uparrow$) & Full($\uparrow$) & Precision($\uparrow$) & Recall($\uparrow$) & F1-Score($\uparrow$) & None($\uparrow$) & Full($\uparrow$) \\ 
            \midrule 
            
            HQ (GT) & 90.03 & 85.52 & 87.72 & 72.06 & 79.48 & 90.29 & 85.77 & 87.97 & 74.50 & 81.72 \\ 
            \cmidrule(lr){1-11} 
            LQ & 89.10 & 44.97 & 59.77 & 42.64 & 50.21 & 85.33 & 51.61 & 64.32 & 47.08 & 55.11 \\ 
            \cmidrule(lr){1-11} 
            Real-ESRGAN~\citep{wang2021real} & 79.15 & 52.70 & 63.27 & 35.30 & 39.88 & {82.67} & 53.94 & 65.29 & 38.16 & 42.36 \\ 
            SwinIR~\citep{liang2021swinir} & 80.29 & 47.45 & 59.64 & 38.39 & 42.63 & {82.92} & 47.89 & 60.72 & 39.97 & 44.56 \\ 
            ResShift~\citep{yue2023resshift} & {81.17} & 33.76 & 47.69 & 30.95 & 34.87 & 82.23 & 39.91 & 53.74 & 35.31 & 39.99 \\ 
            StableSR~\citep{wang2024exploiting} & 79.79 & 59.89 & {68.42} & {41.23} & {47.64} & 82.19 & 60.39 & 69.62 & {42.53} & {49.39} \\ 
            DiffBIR~\citep{lin2024diffbir} & 66.04 & 59.69 & 62.71 & 33.75 & 40.05 & 76.33 & 61.87 & 68.35 & 39.27 & 46.11 \\ 
            SeeSR~\citep{wu2024seesr} & 68.12 & 63.46 & 65.71 & 37.11 & 43.43 & 74.29 & 62.47 & 67.87 & 40.34 & 46.54 \\ 
            SUPIR~\citep{yu2024scaling} & 44.00 & 40.56 & 42.21 & 22.29 & 25.03 & 53.08 & 44.47 & 48.39 & 27.25 & 30.59 \\ 
            FaithDiff~\citep{chen2024faithdiff} & 71.21 & {64.50} & 67.69 & 38.81 & 44.28 & 76.90 & {65.20} & {70.57} & 41.64 & 47.97 \\

            TeReDiff~\cite{min2025text} & \underline{83.95} & \underline{67.58} & \textbf{74.88} & \underline{48.39} & \underline{55.01} & \underline{84.30} & \underline{67.37} & \textbf{74.89} & \underline{49.39} & \underline{56.45} \\ 
            
            DiT4SR~\cite{duan2025dit4sr} & 74.94 & \textbf{68.77} & \underline{71.72} & 46.17 & 51.67 & 79.53& \textbf{69.25} & \underline{74.04} & 48.93 & 54.58\\       
            
            \cmidrule(lr){1-11} 
            \textbf{UniT (Ours)} & \textbf{85.06}	& 60.42	& 70.65 & \textbf{49.81} & \textbf{56.11} & \textbf{86.94}	& 62.74	& 72.88 & \textbf{53.80} & \textbf{59.74}  \\
            \bottomrule    
        \end{tabular}
    }
    \vspace{-7pt}
    \caption{\textbf{Text restoration quantitative results on Real-Text~\cite{min2025text}.} Text restoration performance is evaluated through text detection and recognition by two text spotting modules~\cite{liu2021abcnet, zhang2022text}. ‘None’ and ‘Full’ denote recognition without and with a full lexicon, respectively. Best and second-best results are shown in \textbf{bold} and \underline{underlined}.}
    \label{tab:ts_metric_realtext}    
    \vspace{-5pt}
\end{table*}

\subsection{Inference}
\label{sec_unit:inference_pipeline}

For a denoising timestep sequence $(T-1, \dots, 0)$, the denoising objective at each timestep $t$ is defined as
\begin{equation}
    p_\theta(\mathbf{z}_t \mid \mathbf{z}_{t+1}, \mathbf{z}_{LQ}, \mathbf{z}_{\mathrm{text}}),
\end{equation}
indicating that restoration at each step is conditioned on the previous noisy latent $\mathbf{z}_{t+1}$, the LQ latent $\mathbf{z}_{LQ}$, and the text guidance latent $\mathbf{z}_{\mathrm{text}}$. 

With the TSM integrated into the restoration module, the OCR prediction at timestep $t$ is given by
\begin{equation}
    \mathbf{c}_{\mathrm{tsm},t} \sim p_\psi(\mathbf{c}_{\mathrm{tsm},t} \mid \mathrm{Conv}_{\mathrm{TSM}}(\mathbf{Z}_{\mathrm{cat},t})),
\end{equation}
where $\mathbf{c}_{\mathrm{tsm},t}$ is constructed from the text recognition outputs $\{r_t^{(i)}\}_{i=1}^K$ produced by the recognition decoder $\mathcal{T}_{\mathrm{dec}}^\mathrm{rec}$ of the TSM. 

In cases where the text extracted by the VLM contains errors, we stimulate VLM self-correction by leveraging the OCR predictions generated by the TSM. Specifically, at a predefined correction timestep $j$, the VLM refines its initial text prediction $\mathbf{c}_{\mathrm{vlm}}$ by leveraging the preceding TSM OCR output $\mathbf{c}_{\mathrm{tsm},j+1}$, yielding updated textual guidance $\mathbf{c}_{\mathrm{vlm},j}$, defined as
\begin{equation}
    \mathbf{c}_{\mathrm{vlm},j} \sim p_\phi(\mathbf{c}_{\mathrm{vlm},j} \mid \mathbf{x}_{LQ}, \mathbf{c}_{{\mathrm{tsm}},j+1}).
\end{equation}
This guidance is subsequently encoded into $\mathbf{z}_{\mathrm{vlm},j}$, which serves as conditioning information for the restoration module during the denoising step at timestep $i \leq j-1$ as:
\begin{equation}
    p_\theta(\mathbf{z}_{i} \mid \mathbf{z}_{i+1}, \mathbf{z}_{LQ}, \mathbf{z}_{\mathrm{vlm},j}).
\end{equation}
By leveraging the TSM OCR prediction, the VLM refines its initial prediction and provides more faithful textual guidance, ultimately yielding higher-fidelity text restoration.

After completing the diffusion denoising process, the final text-aware restored image is obtained as
\begin{equation}
    \mathbf{x}_{HQ} = \mathrm{VAE_{Dec}}(\mathbf{z}_{0}),
\end{equation}
where $\mathrm{VAE_{Dec}}$ denotes the VAE decoder~\cite{kingma2013auto}. The predefined correction timesteps, text correction results, and corresponding visualizations are provided in the supplementary material.
\section{Experiments}
\label{sec:4_exp}

\subsection{Experimental Setup}
\label{sec:exp_setup}

\paragraph{Training and evaluation dataset.}
For evaluation, we follow the protocol of TeReDiff~\cite{min2025text} and use the same training and testing setup. The model is trained on SA-Text, which contains 100K high-quality 512$\times$512 images with synthetic degradations generated by the Real-ESRGAN pipeline~\cite{wang2021real}, a standard degradation model in image restoration~\cite{yue2023resshift, wang2024exploiting, lin2024diffbir, wu2024seesr, yu2024scaling}. Testing is performed on SA-Text1K (Lv1-Lv3), the SA-Text test set ($\text{SA-Text}_\text{test}$), and RealText. SA-Text1K provides three levels of progressive degradation for robustness evaluation, while RealText contains real-world HR-LR pairs curated from RealSR~\cite{cai2019toward} and DRealSR~\cite{wei2020component}.
\vspace{-10pt}


\paragraph{Evaluation metrics.}
Text restoration performance is evaluated using pretrained text-spotting models~\cite{liu2021abcnet, zhang2022text}, reporting Precision, Recall, and F1-score for detection, and F1-score for end-to-end recognition. Detection metrics assess the accuracy of text detection, while end-to-end metrics evaluate the correctness of recognized text against ground-truth annotations. Together, these metrics measure functional legibility, capturing both spatial and linguistic fidelity of restored text. This evaluation protocol follows standard practice in text-spotting benchmarks. Image restoration is assessed using reference metrics such as PSNR and SSIM~\cite{wang2004image}, LPIPS~\cite{zhang2018unreasonable}, DISTS~\cite{ding2020image}, and FID~\cite{heusel2017gans}, as well as non-reference metrics including NIQE~\cite{zhang2015feature}, MANIQA~\cite{yang2022maniqa}, MUSIQ~\cite{ke2021musiq}, and CLIPIQA~\cite{wang2023exploring}.

\paragraph{Implementation details.} 
\label{sec:impl}
We employ the DiT backbone from DiT4SR~\cite{duan2025dit4sr} because its integration of the LQ control branch within the transformer blocks provides an effective mechanism for exchanging information between conditional inputs and generative features. For text spotting, we adopt the TESTR architecture~\cite{zhang2022text}, whose design avoids ROI pooling and therefore preserves the fine-grained cues needed to leverage diffusion features. Training is conducted on $512 \times 512$ LQ-HQ image pairs using the AdamW optimizer. Further implementation details are provided in the supplementary material.

\subsection{Comparison with State-of-the-Art}
\label{sec:sota}

\paragraph{Quantitative comparison.}
The text restoration results on SA-Text are shown in Tab.~\ref{tab:metric_ts_satext}, comparing UNet based and DiT based diffusion models. We use the E2E (full) detection and recognition F1 score as the primary measure of text restoration performance. Among UNet based models, TeReDiff~\cite{min2025text} achieves the strongest results, improving accuracy across all degradation levels relative to the original LQ inputs and avoiding hallucinations observed in earlier UNet based approaches. Although Dit4SR~\cite{duan2025dit4sr} is not trained on additional text focused datasets, it still delivers competitive performance, underscoring the value of a strong diffusion backbone for textual reconstruction. Our UniT model further surpasses all baselines by leveraging VLM visual linguistic priors and TSM OCR predictions, achieving the highest restoration performance across all levels of SA-Text. Results on the Real Text dataset in Tab.~\ref{tab:ts_metric_realtext} show similar trends, confirming the robustness of UniT in real world scenarios.

\begin{table}[t]
\centering
\footnotesize
\resizebox{\columnwidth}{!}{
\begin{tabular}{c|c|c|c|c}
    \toprule
        Model & VLM & TSM & E2E(None) F1 ($\uparrow$) & E2E(Full) F1 ($\uparrow$)  \\    
    \midrule
         A & LLaVA (13B) & \ding{55} & 48.14  & 56.24  \\
         B & Qwen2.5-VL (7B) & \ding{55} & \underline{51.21} & \underline{58.94} \\
         C & \ding{55} & \ding{51} & 48.13 & 55.54 \\
         D & Qwen2.5-VL (7B) & \ding{51} & \textbf{53.80}  & \textbf{59.74} \\
    \bottomrule
\end{tabular}}
\vspace{-7pt}
\caption{\textbf{Ablation of VLM and TSM on Real-Text~\cite{min2025text}.} Exploiting stronger VLMs~\cite{liu2023visual, Qwen2.5-VL} consistently improves text restoration performance. Moreover, jointly using the VLM with a text spotting module (TSM)~\cite{zhang2022text} to correct initial VLM text predictions yields the best end-to-end (E2E) accuracy on Real-Text~\cite{min2025text}.}
\vspace{-5pt}
\label{tab:abl_vlm_tsm}
\end{table}

\begin{figure*}[t!]
    \centering
    \includegraphics[width=\linewidth]{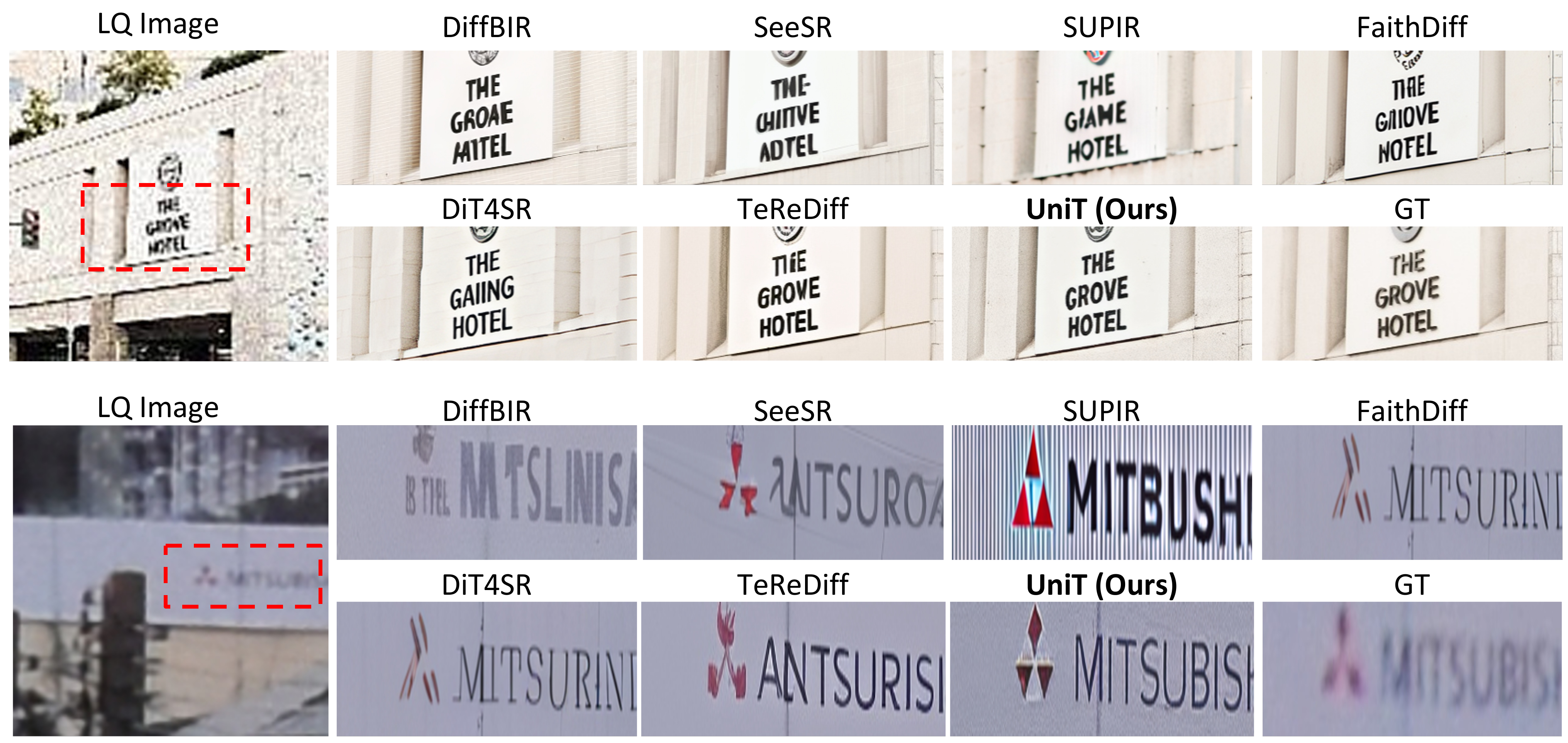}
    \vspace{-20pt}
    \caption{\textbf{Qualitative results on SA-Text and Real-Text benchmarks~\cite{min2025text}.} Compared to the best performing UNet- and DiT-based models \cite{lin2024diffbir, wu2024seesr, yu2024scaling, chen2024faithdiff, duan2025dit4sr, min2025text}, UniT can restore degraded text in images more faithfully to the original. In some cases, UniT can make the text more defined and clear than in the GT, as shown in the bottom figure. Refer to the supplementary material for more visualizations on the three level degradations of SA-Text and on Real-Text benchmarks.}
    \label{fig:qual}
\end{figure*}

\paragraph{Qualitative comparison.}
We present qualitative text restoration results on the SA-Text and Real-Text benchmarks~\cite{min2025text} in Fig.~\ref{fig:qual}. Despite severe degradations affecting readability and style, UniT leverages a rich visual-linguistic prior, guided by precise character-level OCR predictions, to provide accurate textual guidance to the DiT restoration module, effectively recovering the degraded text. In contrast, existing methods frequently fail and often produce hallucinated text. Further qualitative evaluation results for all degradation levels are provided in the supplementary material.

\subsection{Ablation Study}
\label{sec:ablation}

As illustrated in Fig.~\ref{fig:advantage_vlm_tsm}, the VLM and TSM exhibit complementary strengths. By jointly leveraging both models, the restoration framework can receive robust and accurate textual guidance, ultimately producing faithful text restoration results, as shown in Fig.~\ref{fig:complementary_vlm_tsm}. The impact of different VLMs, both with and without the incorporation of the TSM, is presented in Tab.~\ref{tab:abl_vlm_tsm}. Employing a more capable VLM that effectively extracts degraded textual content from LQ images improves restoration accuracy, as observed from model A to model B. Model C, which relies solely on the TSM, exhibits a slight decrease in performance compared to models A and B, which utilize VLMs. However, combining the strongest VLM (model B) with the TSM allows the framework to exploit the complementary characteristics of both models, achieving the highest text restoration performance on the real-world dataset Real-Text~\cite{min2025text}. Refer to the supplementary material for further ablation results.

\section{Conclusion}
\label{sec:5_conclusion}
We propose UniT, a framework for restoring text from severely degraded images by integrating three components: a vision-language model (VLM), a text spotting module (TSM), and a DiT-based restoration backbone. The TSM provides intermediate OCR predictions that enable VLM self-correction, delivering accurate textual guidance to the DiT and achieving state-of-the-art text restoration performance on the SA Text and Real Text benchmarks. However, variability across VLM architectures and prompts suggests that further improvements are possible through adaptive VLM selection, prompt refinement, or tighter VLM-TSM integration. UniT demonstrates the potential of combining diffusion-based restoration with vision-language reasoning for reliable, text-aware image restoration.

\clearpage
\appendix
\section*{\Large Appendix}

This appendix presents analyses, implementation details, and additional results for the UniT framework, organized as follows:

\begin{itemize}
    \setlength{\itemsep}{0.3em} 
    \item Sec.~\ref{supple:sec_A_vlm_lq_extract} presents the query instruction prompts for VLM-based text extraction and outlines the evaluation protocol using an LLM.

    \item Sec.~\ref{supple:sec_B_vlm_self_correct} illustrates how intermediate OCR predictions from the TSM facilitate VLM self-correction.

    \item Sec.~\ref{supple:sec_C_extended_abl} reports extended text restoration results using Null and ground-truth (GT) textual guidance, highlighting the importance of explicit textual cues.

    \item Sec.~\ref{supple:sec_D_unit_framework} describes the UniT architecture, including the restoration and text spotting module (TSM), and details the TSM training strategy.

    \item Sec.~\ref{supple:sec_E_impl_detail} provides additional implementation details regarding the model, datasets, and evaluation metrics.

    \item Sec.~\ref{supple:sec_F_iqa_metric} introduces the image quality assessment (IQA) metrics and investigates their use in quantifying text restoration performance.

    \item Sec.~\ref{supple:sec_G_add_qual} presents extensive qualitative results on Real-Text and SA-Text~\cite{min2025text}, demonstrating UniT’s high-fidelity, text-aware restoration.

    \item Sec.~\ref{supple:sec_H_limitation} discusses the limitations and potential directions for future work in UniT.
\end{itemize}

\section{VLM LQ Text Extraction Details}
\label{supple:sec_A_vlm_lq_extract}

We employ VLMs to extract textual information from LQ images, providing explicit guidance for the restoration model. This section presents LQ text extraction results for different query instructions and the classification methodology for analyzing outputs. Evaluation is conducted on SA-Text Lv3~\cite{min2025text}, consisting of 1,000 severely degraded samples (see Fig.~\ref{fig:suppl_satext_lv_vis}. Our default query, which yielded the best overall performance, is: ``OCR this image and transcribe only the English text.'' We also tested three alternative prompts to assess their effectiveness in extracting degraded text:

\begin{enumerate}
    \item ``Read and transcribe all English text visible in this low-resolution image.''
    \item ``Describe the contents of this blurry image, focusing only on any visible English text or characters.''
    \item ``Extract all visible English words and letters from this low-quality image, even if they appear unclear.''
\end{enumerate}

The results of applying these query instructions to Qwen2.5-VL (7B)~\cite{Qwen2.5-VL}, the best-performing model for LQ text extraction in our study, are provided in Tab.~\ref{tab:query_lq_text_extract}.


\begin{table}[t!]
\centering
\scriptsize 
\setlength{\tabcolsep}{3pt} 
    \begin{tabular}{c|c|c|c|c}
        \toprule
            Question type & Incorrect (\%) & Partial (\%) & Exact (\%) & Partial+Exact (\%)  \\    
        \midrule
            1 &  63.50 & 11.30 & 25.20 & 36.50  \\
            2 &  61.90 & 12.00 & 26.10 &  38.10 \\
            3 &  62.50 & 12.00 & 25.50  & 37.50 \\
        \midrule
            Default & \textbf{60.80} & \textbf{12.40} & \underline{26.80} & \textbf{39.20} \\
        \bottomrule
    \end{tabular}
    \vspace{-5pt}
    \caption{\textbf{LQ text extraction under different query instructions.}
    We report the LQ text extraction performance of Qwen2.5-VL (7B)~\cite{Qwen2.5-VL} across varying OCR query instructions. The outputs are categorized as Incorrect, Partially Correct, or Exactly Correct.}
    \vspace{-10pt}
    \label{tab:query_lq_text_extract}
\end{table}

After extracting textual content from LQ images using the VLM, we evaluate its correctness by comparing it against the ground-truth transcript. Manual verification of over 1,000 samples is both impractical and labor-intensive, thus we employ a large language model (LLM)~\cite{qwen3} to perform this comparison. LLMs possess strong comparative reasoning capabilities and can reliably assess textual correspondence when provided with structured evaluation instructions. Accordingly, we provide the LLM with the VLM-extracted text, the ground-truth transcription, and a prompt specifying the classification criteria as follows:
 
\begin{tcolorbox}[colback=gray!5, colframe=black!15, sharp corners, boxrule=0.4pt, left=6pt, right=6pt, top=6pt, bottom=6pt]
    \ttfamily
    \scriptsize
    \textbf{Ground truth text:} "GT Text"\\
    \textbf{VLM OCR output:} "VLM OCR Output"

    \vspace{2mm}
    \textbf{Evaluation Steps:}
    \begin{enumerate}
        \item Extract the text content from the VLM OCR output.
        \item Compare the extracted text with the ground truth, considering:
        \begin{itemize}
            \item Word order does not matter.
            \item Compare based only on the set of unique words in the ground truth.
            \item Ignore capitalization, punctuation, and extra/missing spaces.
            \item Minor typos still count as matches.
        \end{itemize}
    \end{enumerate}

    \vspace{2mm}
    \textbf{Categories:}
    \begin{enumerate}
        \item Correct: all unique ground truth words appear in the OCR output (ignoring order, case, spacing, typos).
        \item Slightly Correct (Partially Correct): at least one but not all unique words match.
        \item Incorrect: no words match, or the output is largely wrong, unrelated, or empty.
    \end{enumerate}

    \vspace{2mm}
    \textbf{Answer with only the category number (1, 2, or 3).}
\end{tcolorbox}

This LLM-based evaluation offers a scalable approach to assess text extracted by the VLM from low-quality images. By accounting for unique word matches and minor typos while ignoring word order, capitalization, punctuation, and spacing, the LLM classifies outputs as Correct, Partially Correct, or Incorrect. Human validation on 100 samples confirmed the accuracy of this method, making it an effective alternative when human evaluation is impractical.

\section{VLM Self-Correction Analysis}
\label{supple:sec_B_vlm_self_correct}
\begin{figure*}[t]
    \centering
    \includegraphics[width=\linewidth]{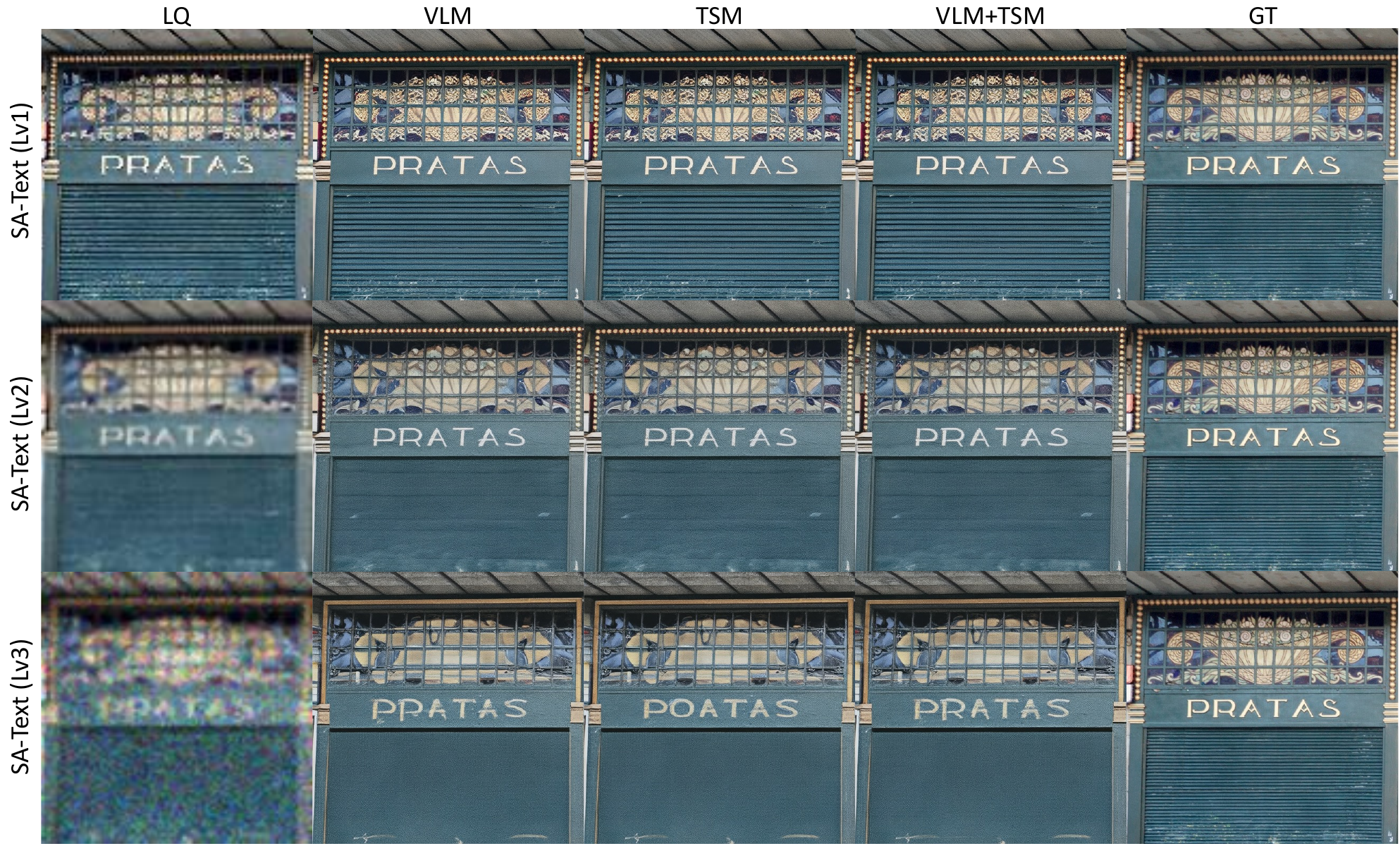}
    \vspace{-20pt}
    
    \caption{\textbf{Text restoration results across three SA-Text degradation levels (Lv1-Lv3).} The figure presents restoration outputs obtained using the VLM alone, the TSM alone, and their combined use (VLM+TSM). While both VLM and TSM yield accurate reconstructions under mild and moderate degradation (Lv1 and Lv2), the severe degradation in Lv3 leads to incorrect predictions from each individual model. In contrast, the joint VLM+TSM configuration successfully restores the ground-truth text “PRATAS”.}

    \label{fig:suppl_satext_lv_vis}
    \vspace{-5pt}
\end{figure*}

VLMs leverage strong visual-linguistic priors to infer degraded textual content from LQ images. However, their performance diminishes when the degraded text is semantically inconsistent with the scene or when visual cues are insufficient to activate these priors. In such scenarios, intermediate OCR predictions generated by the TSM can facilitate VLM self-correction. By jointly considering the TSM's OCR predictions and the LQ image, the VLM can rectify initial errors, thereby providing more reliable textual guidance for the downstream restoration model. The raw OCR predictions and corrected results of the TSM and VLM of Fig.~\ref{fig:complementary_vlm_tsm} (c) is presented in Fig.~\ref{fig:vlm_raw_self_correction}. Moreover, under severe degradation conditions such as the SA-Text Lv3 setting, using either the VLM or the TSM in isolation produces incorrect textual guidance and consequently erroneous restorations, whereas the combined VLM+TSM approach yields accurate text restoration, as shown in Fig.~\ref{fig:suppl_satext_lv_vis}.

\begin{figure}[t!]
\centering
\begin{tcolorbox}[colback=gray!5, colframe=black!15, sharp corners, boxrule=0.4pt, left=6pt, right=6pt, top=6pt, bottom=6pt]
\ttfamily
\scriptsize

\begin{tabular}{@{}l@{}}
\cellcolor{yellow}[VLM initial text prompt]: 104
\end{tabular} \\[3pt]

\begin{tabular}{@{}l l@{}}
iter: 00  & TSM OCR text prompt: [200] \\[1pt]
iter: 01  & TSM OCR text prompt: [2.0] \\[1pt]
iter: 02  & TSM OCR text prompt: [210] \\[1pt]
iter: 03  & TSM OCR text prompt: [210] \\[1pt]
iter: 04  & TSM OCR text prompt: [210] \\[1pt]
iter: 05  & TSM OCR text prompt: [210] \\[1pt]
iter: 06  & TSM OCR text prompt: [219] \\[1pt]
iter: 07  & TSM OCR text prompt: [220] \\[1pt]
iter: 08  & TSM OCR text prompt: [210] \\[1pt]
iter: 09  & TSM OCR text prompt: [210] \\[3pt]

\rowcolor{yellow}
iter: 10  & APPLY VLM CORRECTION: [210] \\[1pt]
iter: 11  & vlm corrected prompt: [210] \\[1pt]
iter: 12  & vlm corrected prompt: [210] \\[1pt]
iter: 13  & vlm corrected prompt: [210] \\[1pt]
iter: 14  & vlm corrected prompt: [210] \\[1pt]
iter: 15  & vlm corrected prompt: [210] \\[1pt]
iter: 16  & vlm corrected prompt: [210] \\[1pt]
iter: 17  & vlm corrected prompt: [210] \\[1pt]
iter: 18  & vlm corrected prompt: [210] \\[1pt]
iter: 19  & vlm corrected prompt: [210] \\[3pt]

\rowcolor{yellow}
iter: 20  & APPLY VLM CORRECTION: [211] \\[1pt]
iter: 21  & vlm corrected prompt: [211] \\[1pt]
iter: 22  & vlm corrected prompt: [211] \\[1pt]
iter: 23  & vlm corrected prompt: [211] \\[1pt]
iter: 24  & vlm corrected prompt: [211] \\[1pt]
iter: 25  & vlm corrected prompt: [211] \\[1pt]
iter: 26  & vlm corrected prompt: [211] \\[1pt]
iter: 27  & vlm corrected prompt: [211] \\[1pt]
iter: 28  & vlm corrected prompt: [211] \\[1pt]
iter: 29  & vlm corrected prompt: [211] \\[1pt]
iter: 30  & vlm corrected prompt: [211] \\[1pt]
iter: 31  & vlm corrected prompt: [211] \\[1pt]
iter: 32  & vlm corrected prompt: [211] \\[1pt]
iter: 33  & vlm corrected prompt: [211] \\[1pt]
iter: 34  & vlm corrected prompt: [211] \\[1pt]
iter: 35  & vlm corrected prompt: [211] \\[1pt]
iter: 36  & vlm corrected prompt: [211] \\[1pt]
iter: 37  & vlm corrected prompt: [211] \\[1pt]
iter: 38  & vlm corrected prompt: [211] \\[1pt]
iter: 39  & vlm corrected prompt: [211]
\end{tabular}

\end{tcolorbox}
\vspace{-10pt}
\caption{\textbf{Raw VLM self-correction results for Fig.~\ref{fig:complementary_vlm_tsm} (c).} VLM self-correction is performed at denoising iteration steps 10 and 20. The initial VLM LQ text extraction result, "104", serves as guidance for the restoration model until iteration 10. At iteration 10, the VLM leverages the TSM OCR prediction from iteration 9 (210) to correct its initial prediction to "210". At iteration 20, the VLM further refines its prediction to "211", successfully correcting its initial prediction. Rows corresponding to VLM corrections are highlighted in yellow.}

\label{fig:vlm_raw_self_correction}
\end{figure}

\section{Extended Ablation Experiment}
\label{supple:sec_C_extended_abl}
\begin{figure}[t]
    \centering
    \includegraphics[width=\linewidth]{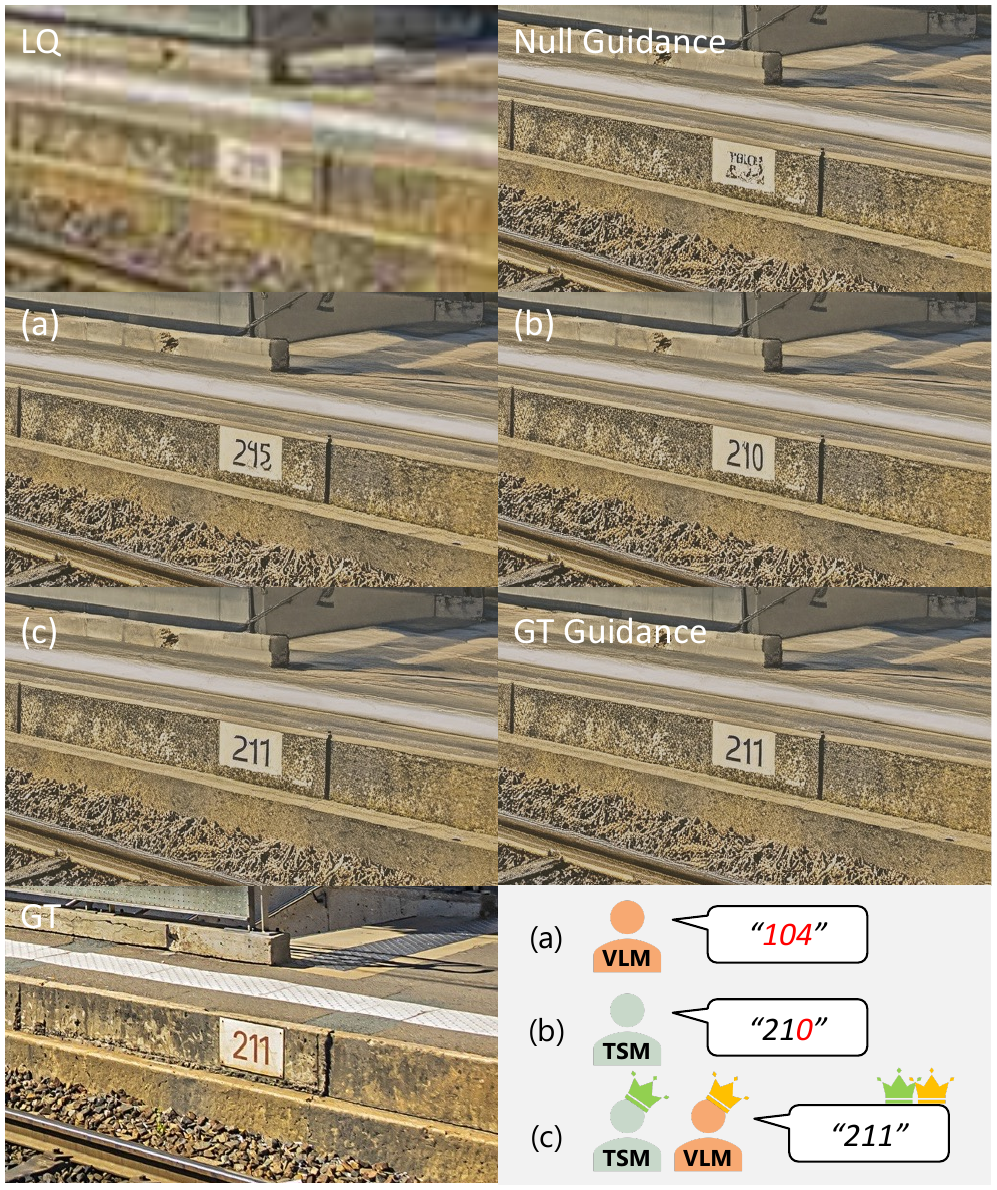}
    \vspace{-20pt}
    
    \caption{\textbf{Full ablation including Null and GT guidance.} Extended text restoration results from Fig.~\ref{fig:complementary_vlm_tsm}, incorporating Null and ground-truth (GT) guidance settings on Real-Text~\cite{min2025text}.}
    
    \label{fig:supple_full}
    \vspace{-10pt}
\end{figure}

\begin{figure}[t]
    \centering
    \includegraphics[width=\linewidth]{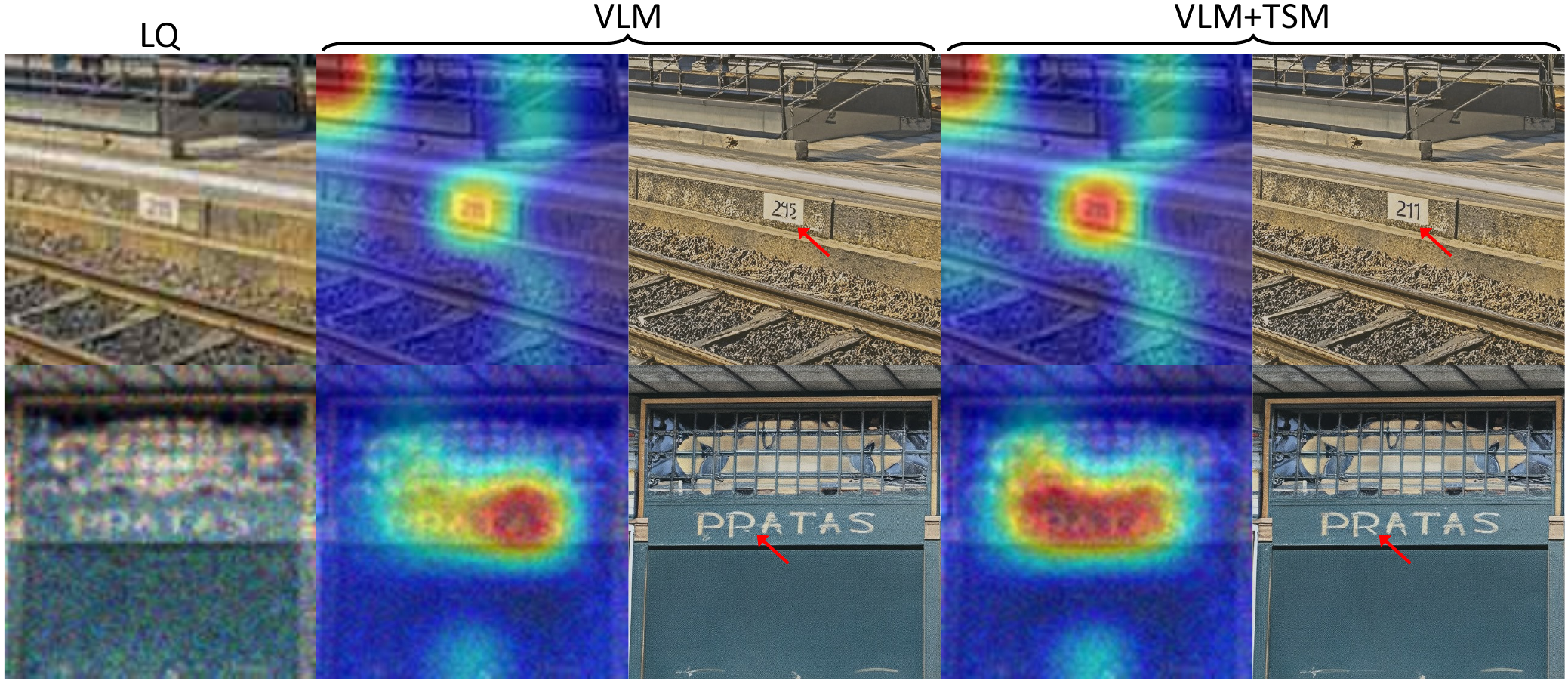}
    \vspace{-20pt}
    \caption{\textbf{VLM~\cite{Qwen2.5-VL} attention map visualization.} The intermediate OCR predictions generated by TSM function as reliable textual grounding for correcting erroneous VLM~\cite{Qwen2.5-VL} text predictions, thereby enabling the VLM to refine its initial textual output and achieve more accurate text restoration.}
    \label{fig:attn_map}
    \vspace{-10pt}
\end{figure}

In addition to Tab.~\ref{tab:abl_vlm_tsm} and Fig.~\ref{fig:complementary_vlm_tsm} in the main paper, Tab.~\ref{tab:full_null_vlm_tsm_gt}, Fig.~\ref{fig:supple_full}, and Fig.~\ref{fig:attn_map} present the full performance results, including evaluations with Null prompts, ground-truth textual guidance, and attention-map analysis. These ablation studies allow us to assess the effect of providing explicit textual cues to the restoration model. Incorporating such cues improves text-restoration performance over the baseline, while supplying ground-truth guidance yields an additional substantial gain, representing an idealized upper bound. Although ground-truth guidance is not feasible in real-world scenarios, these findings highlight the potential benefits of leveraging accurate textual cues to enhance restoration quality.

\begin{table}[t]
    \centering
    \footnotesize
    \resizebox{0.95\columnwidth}{!}{
        \begin{tabular}{c|c|c|c|c}
            \toprule
                Model & VLM & TSM & E2E(None) F1 ($\uparrow$) & E2E(Full) F1 ($\uparrow$)  \\    
            \midrule
                 Null Guidance & \ding{55} & \ding{55} & 48.04 & 54.51 \\
            \midrule
                 A & LLaVA (13B) & \ding{55} & 48.14  & 56.24  \\
                 B & Qwen2.5-VL (7B) & \ding{55} & \underline{51.21} & \underline{58.94} \\
                 C & \ding{55} & \ding{51} & 48.13 & 55.54 \\
                 D & Qwen2.5-VL (7B) & \ding{51} & \textbf{53.80}  & \textbf{59.74} \\
            \midrule
                 GT Guidance & \ding{55} & \ding{55} & 59.66 & 66.73 \\
            \bottomrule
        \end{tabular}}
        
    \vspace{-5pt}
    \caption{\textbf{Full ablation including Null and GT guidance.} Extended text restoration performance from Tab.~\ref{tab:abl_vlm_tsm}, incorporating Null and GT guidance settings on Real-Text~\cite{min2025text}. Providing guidance to the restoration model improves text restoration performance, while GT guidance yields a substantial additional gain. However, GT guidance is not feasible in practice, as it represents an idealized scenario.}
    \vspace{-5pt}
    \label{tab:full_null_vlm_tsm_gt}
\end{table}

\section{UniT Framework Details}
\label{supple:sec_D_unit_framework}

\subsection{Architecture Details}
The architectures of the DiT-based image restoration module, vision-language model, and the text spotting module are based on DiT4SR~\cite{duan2025dit4sr} Qwen2.5-VL~\cite{Qwen2.5-VL} and TESTR~\cite{zhang2022text}, respectively, each initialized with publicly available checkpoints. 

Given a LQ image $\mathbf{x}_{LQ} \in \mathbb{R}^{H \times W \times 3}$, the objective is to restore a HQ image $\mathbf{x}_{HQ} \in \mathbb{R}^{H \times W \times 3}$. Following DiT4SR~\cite{duan2025dit4sr}, which builds on SD3~\cite{esser2024scaling}, an additional LQ branch is incorporated into the MM-DiT block. Thus, alongside the noisy latent $\textbf{z}_t$ and text latent $\textbf{z}_{\mathrm{text}} \in \mathbb{R}^{M \times D}$, the block also receives the LQ latent $\textbf{z}_{LQ}$. Specifically, for denoising timestep $t$, $\textbf{x}_{LQ}$ is encoded with a VAE~\cite{kingma2013auto} to produce a $\times 8$ downsampled latent representation. This representation is then patch-embedded (patch size 2) and augmented with positional encoding~\cite{esser2024scaling}, yielding both the LQ latent $\textbf{z}_{LQ} \in \mathbb{R}^{N \times D}$ and the noisy latent $\textbf{z}_t \in \mathbb{R}^{N \times D}$, where $H=512$, $W=512$, $N=1024$, $M=154$ and $D=1536$. For simplicity, we omit detailed drawing of the timestep embedding and pooled text latent as it follows prior works~\cite{esser2024scaling, duan2025dit4sr}. The three input latents ($\textbf{z}_t$, $\textbf{z}_{LQ}$, and $\textbf{z}_{\mathrm{text}}$) are processed by ${N} = 24$ DiT blocks, each consisting of two main operations: a full-attention layer $\mathrm{Attn}^{\mathrm{n}}$ and a feedforward layer $\mathrm{FF}^{\mathrm{n}}$. The features provided to the TSM are the stacked DiT features $\mathrm{\textbf{Z}}_{\mathrm{cat},t} = (\textbf{z}_{\mathrm{cat},t}^1, \textbf{z}_{\mathrm{cat},t}^2, \ldots, \textbf{z}_{\mathrm{cat},t}^{\mathrm{N}})$, where each $\textbf{z}_{\mathrm{cat},t}^i \in \mathbb{R}^{2L \times D}$ is constructed by concatenating the feedforward outputs $\textbf{z}_{LQ}^i$ and $\textbf{z}_t^i$, both in $\mathbb{R}^{L \times D}$. Lastly, a convolution operation $\mathrm{Conv}_{\mathrm{TSM}}(\cdot)$ is applied to the stacked diffusion features $\mathrm{\textbf{Z}}_{\mathrm{cat},t}$, and the resulting features are passed to the TSM, composed of one encoder $\mathcal{T}_{\mathrm{enc}}$ and two decoders $\mathcal{T}_{\mathrm{dec}}^d$ and $\mathcal{T}_{\mathrm{dec}}^k$. The TSM outputs text detection and recognition results $\{\textbf{d}_t^{(i)}\}_{i=1}^K$ and $\{\textbf{r}_t^{(i)}\}_{i=1}^K$, respectively, where $K$ denotes the number of detected text instances with confidence scores above a threshold $T = 0.5$. Each detection $\textbf{d}^{(i)} = (\textbf{d}_1^{(i)}, \ldots, \textbf{d}_{U}^{(i)})$ is represented as a polygon with $U = 16$ control points, and each recognition result $\textbf{r}^{(i)} = (\textbf{r}_1^{(i)}, \ldots, \textbf{r}_V^{(i)})$ contains $V = 25$ predicted characters.

\subsection{Training Details}
We follow the TSM training protocol described in TeReDiff~\cite{min2025text}, which employs a three-stage training strategy. In the first stage, only the image restoration module is trained. In the second stage, only the TSM is trained. In the third stage, both modules are trained jointly using a combined optimization objective. We observe that while joint training benefits the TSM, it negatively affects the diffusion features of the restoration module. Consequently, we adopt a two-stage training strategy for training the TSM illustrated in Fig.~\ref{fig:framework} (b).

\paragraph{Stage 1 training.}
In Stage~1, only the diffusion-based image restoration module is trained. The module predicts a noise residual 
$\boldsymbol{\epsilon}_t = \hat{\boldsymbol{\epsilon}}_\theta(\mathbf{z}_t, \mathbf{z}_{LQ}, \mathbf{z}_{\mathrm{text}}, t)$
given the noisy latent $\mathbf{z}_t$, the LQ latent $\mathbf{z}_{LQ}$, the text guidance latent $\mathbf{z}_{\mathrm{text}}$, and the denoising timestep $t$.
Following the flow-matching formulation in SD3~\cite{esser2024scaling}, the noisy latent is computed as
\begin{equation}
    \mathbf{z}_t = (1 - \sigma_t)\mathbf{z}_0 + \sigma_t \boldsymbol{\epsilon}, 
    \quad \boldsymbol{\epsilon} \sim \mathcal{N}(\mathbf{0}, \mathbf{I}),
\end{equation}
where $\sigma_t$ denotes the noise scale at timestep $t$, and $t$ is sampled from the noise scheduler $t \sim p(t)$, e.g., a logit-normal distribution.
The model is trained by minimizing the weighted mean-squared error between the predicted noise and the true noise:
\begin{equation}
    \mathcal{L}_{\mathrm{diff}}
    =
    \mathbb{E}_{t, \mathbf{z}_0, \boldsymbol{\epsilon}}
    \Big[
        w_t \,
        \|
        \boldsymbol{\epsilon} - 
        \hat{\boldsymbol{\epsilon}}_\theta(\mathbf{z}_t, \mathbf{z}_{LQ}, \mathbf{z}_{\mathrm{text}}, t)
        \|_2^2
    \Big],
\end{equation}
where $w_t$ is a timestep-dependent weighting factor determined by the SD3 noise schedule. This objective encourages the image restoration module to accurately model the denoising process while leveraging both LQ and textual guidance.

\paragraph{Stage 2 training.}
In Stage2, only the TSM is trained, while the image restoration module remains frozen to provide stable diffusion features.
We employ the same loss formulation as in transformer-based text-spotting methods~\citep{huang2024bridging, zhang2022text, qiao2024dntextspotter}, in which detection and recognition losses are computed via bipartite matching~\citep{carion2020end} and solved using the Hungarian algorithm~\citep{kuhn1955hungarian}. 
Specifically, separate loss functions are applied to the encoder and the dual decoder:
\begin{align}
\mathcal{L}_{\mathrm{enc}} &= 
\sum_m \Big(
    \lambda_{\mathrm{cls}} \mathcal{L}^{(m)}_{\mathrm{cls}}
    + \lambda_{\mathrm{box}} \mathcal{L}^{(m)}_{\mathrm{box}}
    + \lambda_{\mathrm{gIoU}} \mathcal{L}^{(m)}_{\mathrm{gIoU}}
\Big), \label{eq:ts_enc} \\
\mathcal{L}_{\mathrm{dec}} &= 
\sum_n \Big(
    \lambda_{\mathrm{cls}} \mathcal{L}^{(n)}_{\mathrm{cls}}
    + \lambda_{\mathrm{poly}} \mathcal{L}^{(n)}_{\mathrm{poly}}
    + \lambda_{\mathrm{char}} \mathcal{L}^{(n)}_{\mathrm{char}}
\Big), \label{eq:ts_dec}
\end{align}
where $m$ and $n$ index the set of instances with confidence scores exceeding a predefined threshold $T=0.5$.
Here, $\mathcal{L}_{\mathrm{cls}}$, $\mathcal{L}_{\mathrm{box}}$, $\mathcal{L}_{\mathrm{gIoU}}$, $\mathcal{L}_{\mathrm{poly}}$, and $\mathcal{L}_{\mathrm{char}}$ denote the text classification loss, bounding box regression loss, generalized IoU loss~\citep{rezatofighi2019generalized}, polygon regression loss (L1), and character recognition loss (cross-entropy), respectively.
Each term is weighted by its corresponding factor: $\lambda_{\mathrm{cls}}=2.0$, $\lambda_{\mathrm{box}}=5.0$, $\lambda_{\mathrm{gIoU}}=2.0$, $\lambda_{\mathrm{poly}}=5.0$, and $\lambda_{\mathrm{char}}=4.0$.
The final loss function used for the TSM is:
\begin{equation}
    \mathcal{L}_{\mathrm{TSM}} = \mathcal{L}_{\mathrm{enc}} + \mathcal{L}_{\mathrm{dec}}.
\end{equation}

\subsection{Inference Details}


\paragraph{VLM self-correction.}
During denoising, the TSM produces OCR predictions at each timestep. Selecting an appropriate correction timestep for the VLM is critical, as it determines when the VLM uses the TSM output to refine its initial prediction. Early timesteps (e.g., 1-5) yield unstable TSM outputs, which can degrade VLM performance. Optimal performance is observed at timesteps 10-20, allowing the VLM to effectively leverage TSM predictions for self-correction. Corrections applied near the end of denoising (e.g., 30-40) have limited impact on the final restored text, as the denoising process is nearly complete. In all experiments, VLM correction is applied at timestep 10, or at both timesteps 10 and 20.


\begin{table*}[t]
    \centering
    \resizebox{\textwidth}{!}{
        \begin{tabular}{ l|l|cccccccc } 
            \toprule
            \addlinespace[4pt]
            Dataset &
            Model &
            PSNR($\uparrow$) &
            SSIM($\uparrow$) &
            LPIPS($\downarrow$) &
            DISTS($\downarrow$) &
            NIQE($\downarrow$) &
            MANIQA($\uparrow$) &
            MUSIQ($\uparrow$) &
            CLIPIQA($\uparrow$)
            \\
            \addlinespace[3pt]
            \midrule 

            \multirow{4}{*}{$\text{SA-Text}_\text{test}$} 
            & TeReDiff~\cite{min2025text} & \textbf{19.71}	&  \textbf{0.5603}	&  \textbf{0.2828}	&  \textbf{0.1702}	&  5.450	&  \textbf{0.6036}	& 72.06	 &  \underline{0.7023} \\
            & DreamClear~\cite{ai2024dreamclear} & \underline{19.09}	& \underline{0.4978}	& 0.3562	& 0.2109	& \textbf{4.527}	& 0.5545	& 65.69	& 0.6056 \\ 
            & DiT4SR~\cite{duan2025dit4sr} & 18.22 &	0.4900 &	\underline{0.3499} &	\underline{0.1994} &	\underline{5.148} &	0.5225 &	\textbf{72.86} &	{0.6726 }\\ 
            \cmidrule{2-10}
            & \textbf{UniT (Ours)} & 18.72	& 0.4762	& 0.3525	& 0.2070	& 6.328	& \underline{0.5970}	& \underline{72.69} &	\textbf{0.7239} \\
            \midrule 

            \multirow{4}{*}{Real-Text} 
            & TeReDiff~\cite{min2025text} & \underline{23.00} & \textbf{0.7764} & \textbf{0.2848} & 0.2386 & \underline{7.639} & 0.5019 & 62.02 & \underline{0.6441} \\ 
            & DreamClear~\cite{ai2024dreamclear} & \textbf{24.34} & \underline{0.7749} & \underline{0.2936} & \textbf{0.2328} & \textbf{6.848} & 0.3929 & 49.09 & 0.5066 \\ 
            & DiT4SR~\cite{duan2025dit4sr} & 22.75 & 0.7352 & 0.3079 & \underline{0.2342} & 7.752 & \underline{0.4651} & \textbf{62.83} & 0.6370 \\  
            \cmidrule{2-10}
            & \textbf{UniT (Ours)} & 21.84	& 0.5648	& 0.4657	& 0.2841	& 8.463	& \textbf{0.5373}	& \underline{63.62}	& \textbf{0.6976} \\
            \bottomrule    
        \end{tabular}
    }
    \vspace{-4pt}
    \caption{\textbf{Image restoration quantitative results.} Performance comparison on {SA-Text}$_\text{test}$ and {Real-Text} benchmarks~\cite{min2025text}.}
    \label{tab:metric_ir}

\end{table*}

\section{Implementation Details}
\label{supple:sec_E_impl_detail}
\paragraph{Model details.}
The UniT text restoration framework comprises three components: a DiT-based restoration model, a VLM, and a TSM. The DiT architecture follows DiT4SR~\cite{duan2025dit4sr}, the VLM is instantiated using Qwen2.5-VL (7B)~\cite{Qwen2.5-VL}, and the TSM adopts the TESTR design~\cite{zhang2022text}. All modules are initialized with publicly released pretrained weights. The DiT contains 2716.81M parameters, the TSM contains 47.40M parameters, and Qwen2.5-VL contains 7B parameters. During Stage~1, we train the DiT-based image restoration model by unfreezing all parameters in the LQ branch of the MM-DiT Control~\cite{duan2025dit4sr} and the attention layers (Q, K, V projections) in the text and noise branches. Training is performed for five epochs with a learning rate of $5 \times 10^{-5}$ and an effective batch size of 64 obtained through gradient accumulation. The total number of trainable parameters in this stage is 1047.32M. In Stage~2, all other modules are frozen, and only the TSM is fine-tuned. Training is conducted for 10 epochs with a learning rate of $1 \times 10^{-5}$ and an effective batch size of 64. The total number of trainable parameters in this stage is 47.40M. All experiments were conducted on NVIDIA RTX 5880 GPUs.
\vspace{-5pt}

\begin{figure}[!t]
    \centering
    \includegraphics[width=\linewidth]{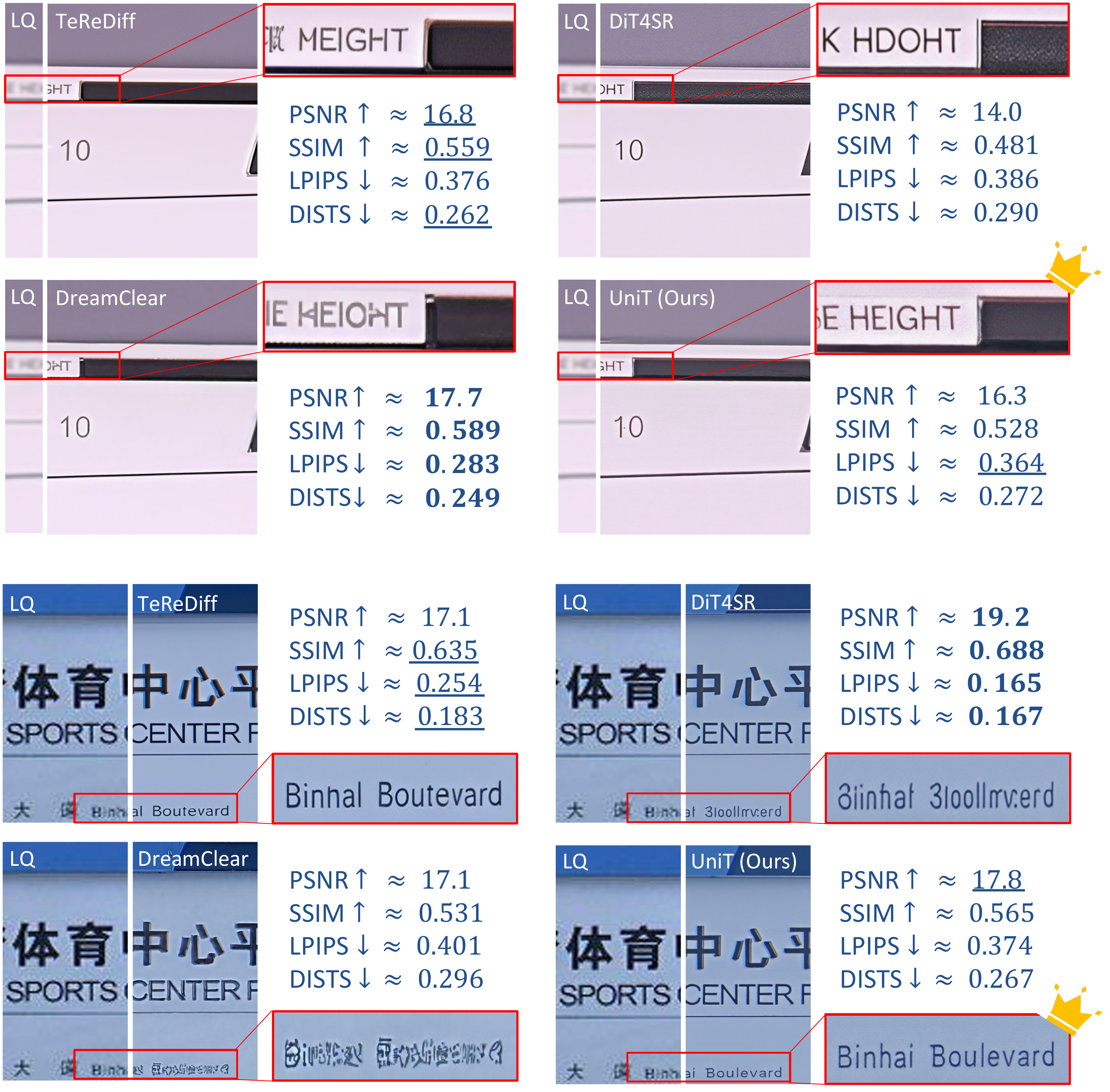}
    \vspace{-20pt}
    \caption{\textbf{Text restoration and cropped FR metric analysis.} Visualization of restored text regions with full-reference (FR) metric scores computed on the cropped regions, illustrating the mismatch between text perceptual quality and reference-based metrics. The best score is in \textbf{bold} and the second best score is \underline{underlined}.}
    \label{fig:qual_metrics}
    \vspace{-15pt}
\end{figure}
\paragraph{Dataset details.}
We use the SA-Text~\cite{min2025text} training dataset, which consists of 100K high-quality $512 \times 512$ images containing text detection and text recognition annotations, covering diverse text fonts, sizes, orientations, and complex visual content. For evaluation, we follow the protocol established in TeReDiff~\cite{min2025text}, where a 1K testing subset of SA-Text is curated and processed using the Real-ESRGAN pipeline~\cite{wang2021real} to obtain SA-Text$_\text{test}$. However, since small text regions are highly sensitive to degradation (i.e., even slight degradation can render them unreadable), three levels of degradation are applied by controlling the kernel strength of the Real-ESRGAN pipeline, resulting in three additional datasets corresponding to different degradation levels: SA-Text (lv1, lv2, lv3). In addition to the synthetically degraded images, we also evaluate on Real-Text~\cite{min2025text}, which consists of real-world LQ-HQ pairs extracted from RealSR~\cite{cai2019toward} and DRealSR~\cite{wei2020component}, allowing us to assess model performance in practical real-world scenarios.
\vspace{-5pt}

\paragraph{Metric details.}
We evaluate text restoration performance following TeReDiff~\cite{min2025text}, which employs text-spotting modules~\citep{liu2020abcnet, liu2021abcnet, zhang2022text, huang2024bridging} for assessment, as accurate reconstruction of degraded textual content is effectively captured by the dedicated text spotting module~\cite{liu2021abcnet, zhang2022text}. In particular, we report Precision, Recall, and F1-score and consider a detection correct if its Intersection over Union (IoU) with a ground-truth bounding box exceeds 0.5. For recognition, we adopt two lexicon-based evaluation settings: None and Full. The None setting evaluates recognition without any lexicon, requiring exact matches to ground-truth transcriptions and reflecting open-vocabulary performance. The Full setting permits predictions to match the closest entry in a ground-truth lexicon, simulating closed-vocabulary conditions. This dual evaluation framework provides a comprehensive measure of recognition accuracy.
\section{Image Quality Metric}
\label{supple:sec_F_iqa_metric}

We evaluate UniT on image quality assessment (IQA) metrics consisting of both FR and NR image quality metrics, as reported in Tab.~\ref{tab:metric_ir}. FR metrics (PSNR, SSIM, LPIPS, DISTS)~\cite{wang2004image, zhang2018unreasonable, ding2020image} measure pixel- or feature-level similarity to the GT, whereas NR metrics (NIQE, MANIQA, MUSIQ, CLIPIQA)~\cite{zhang2015feature, yang2022maniqa, ke2021musiq, wang2023exploring} evaluate perceptual naturalness and semantic quality without requiring GT references.

We compare UniT to the leading text restoration UNet model, TeReDiff~\cite{min2025text}, as well as to DiT-based restoration models~\cite{ai2024dreamclear,duan2025dit4sr}. UniT demonstrates strong performance on NR metrics, indicating improved perceptual fidelity and semantic consistency, but shows comparatively weaker performance on FR metrics (Tab.~\ref{tab:metric_ir}). This discrepancy arises for two reasons. First, FR metrics are sensitive to minor pixel-level deviations; enhancements in character sharpness and structural clarity introduce differences from the GT image, which FR metrics penalize despite visible improvements. Second, the GT text images themselves are not sharply rendered, and restoring them to a clearer form increases the discrepancy between the enhanced output and the inherently soft GT, further lowering FR scores (see Fig.~\ref{fig:qual_realtext}-\ref{fig:qual_satext_3}).

To examine this effect, we visualize the restored results and additionally compute FR metrics on cropped text regions in Fig.~\ref{fig:qual_metrics}. Even when the restored text is structurally accurate, FR scores remain low, underscoring the misalignment between FR metrics and perceptual quality in text-focused restoration. Overall, these findings show that while UniT may not maximize FR metrics, the qualitative evidence and region-level analysis demonstrate its strong restoration capability. For text-image restoration, perceptual and task-oriented metrics provide a more reliable assessment than strict pixel-level comparisons, particularly when GT images contain soft or low-quality text.

\section{More Text Restoration Qualitative Results}
\label{supple:sec_G_add_qual}
\begin{figure*}[!ht]
    \centering
    \includegraphics[width=\linewidth]{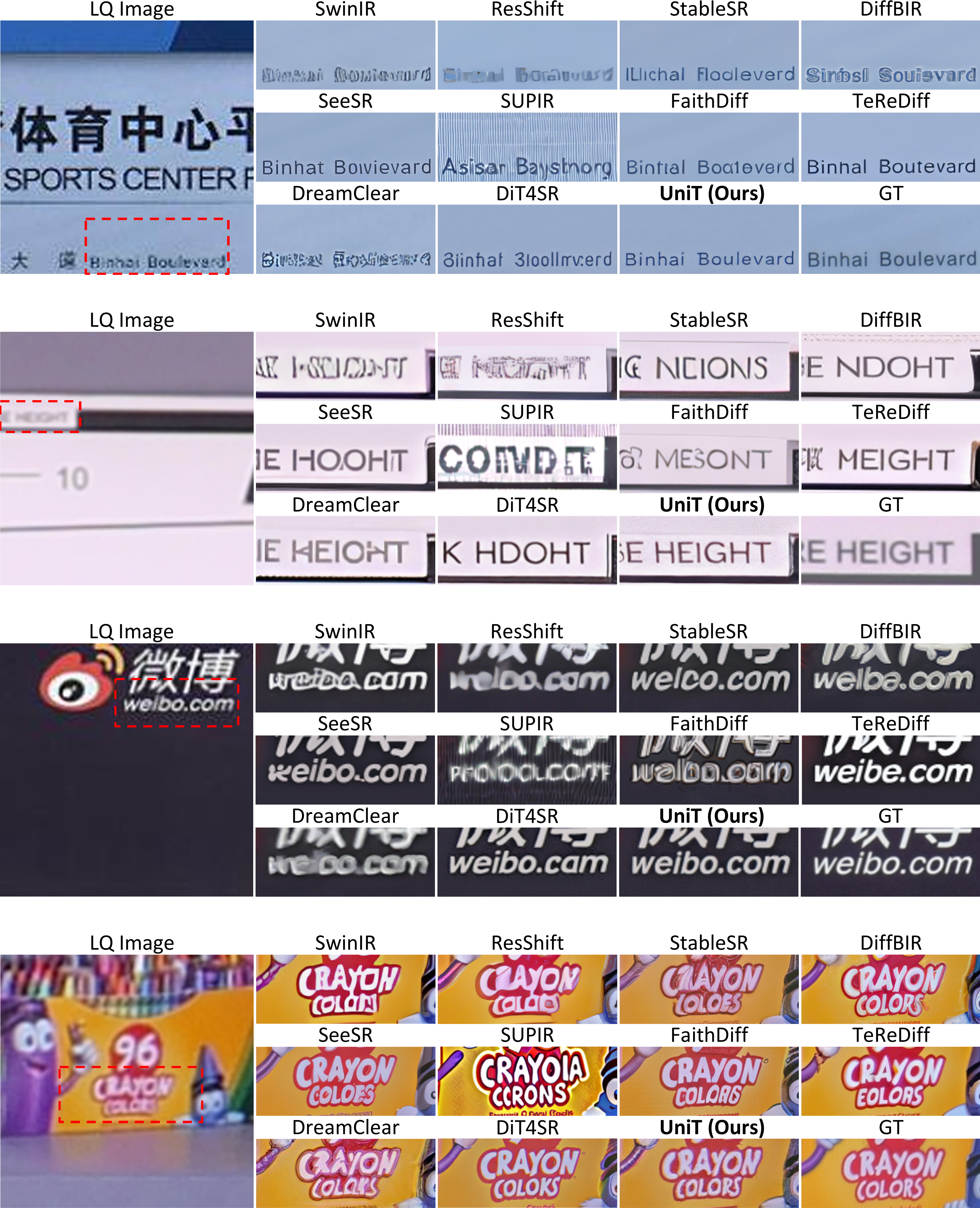}
    \caption{\textbf{Text restoration results on Real-Text~\cite{min2025text}.}}
    \label{fig:qual_realtext}
\end{figure*}

\begin{figure*}[h!]
    \centering
    \includegraphics[width=\linewidth]{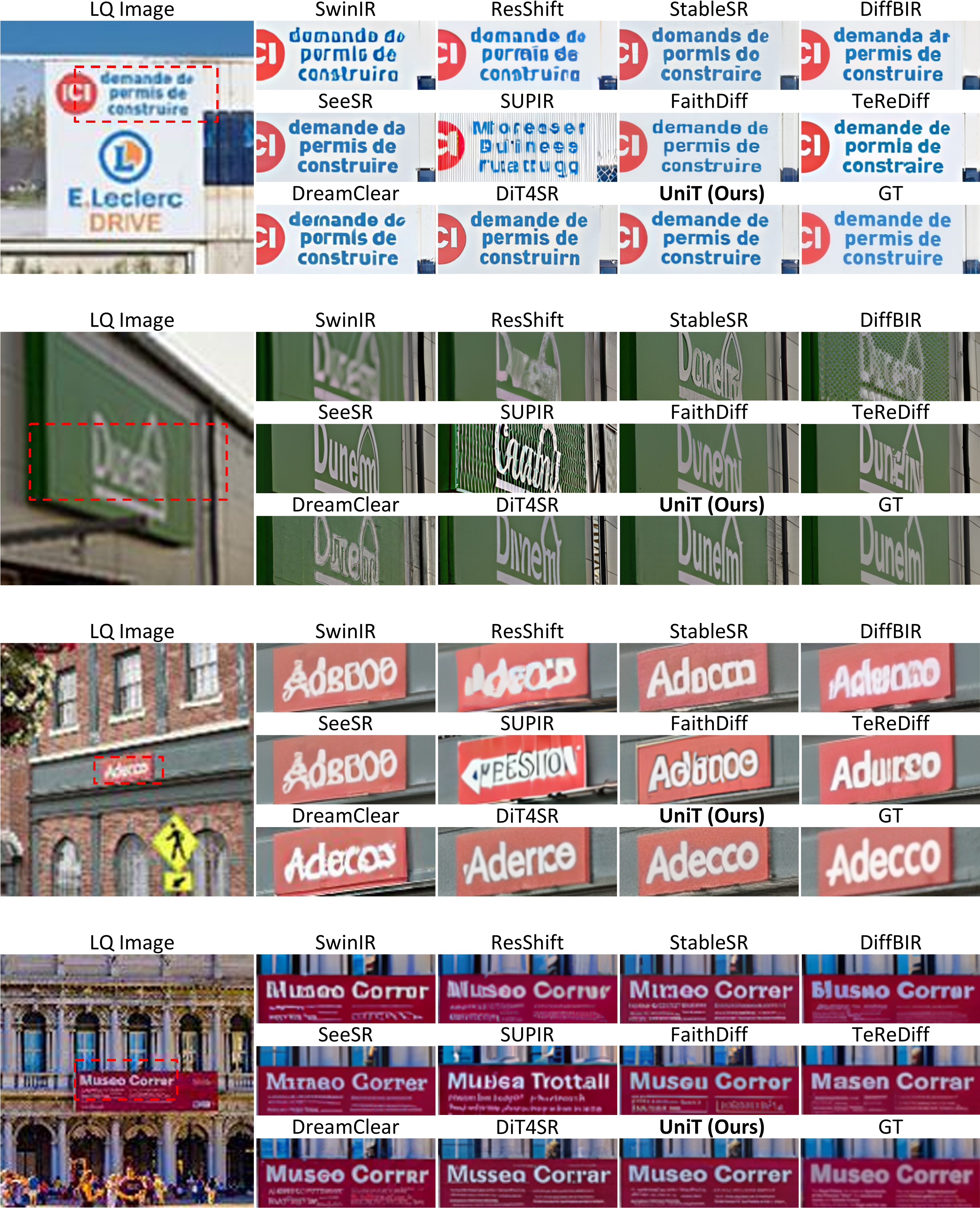}
    \caption{\textbf{Text restoration results on SA-Text test set Level 1.}}
    \label{fig:qual_satext_1}
\end{figure*}

\begin{figure*}[h!]
    \centering
    \includegraphics[width=\linewidth]{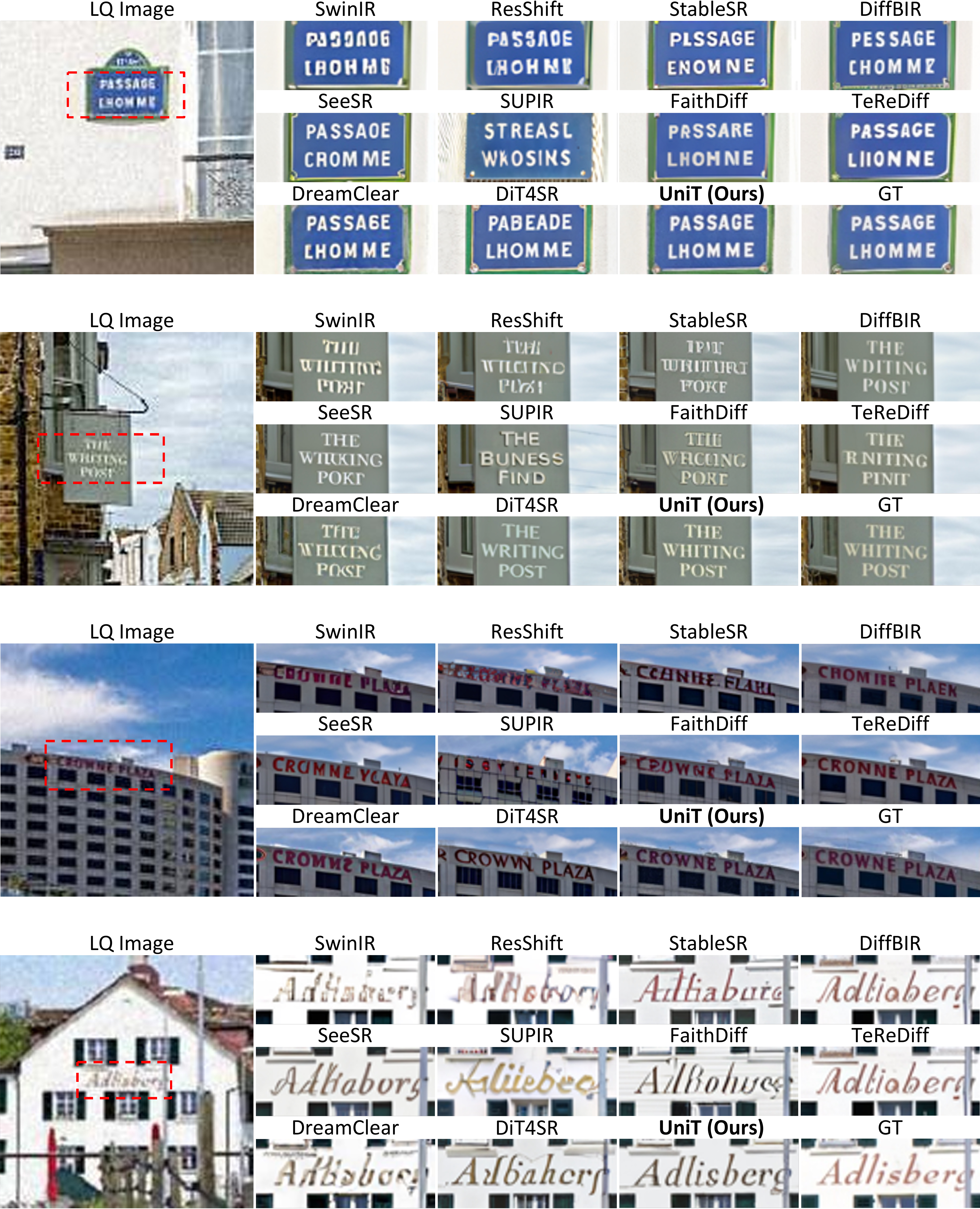}
    \caption{\textbf{Text restoration results on SA-Text test set Level 2.}}
    \label{fig:qual_satext_2}
\end{figure*}

\begin{figure*}[h!]
    \centering
    \includegraphics[width=\linewidth]{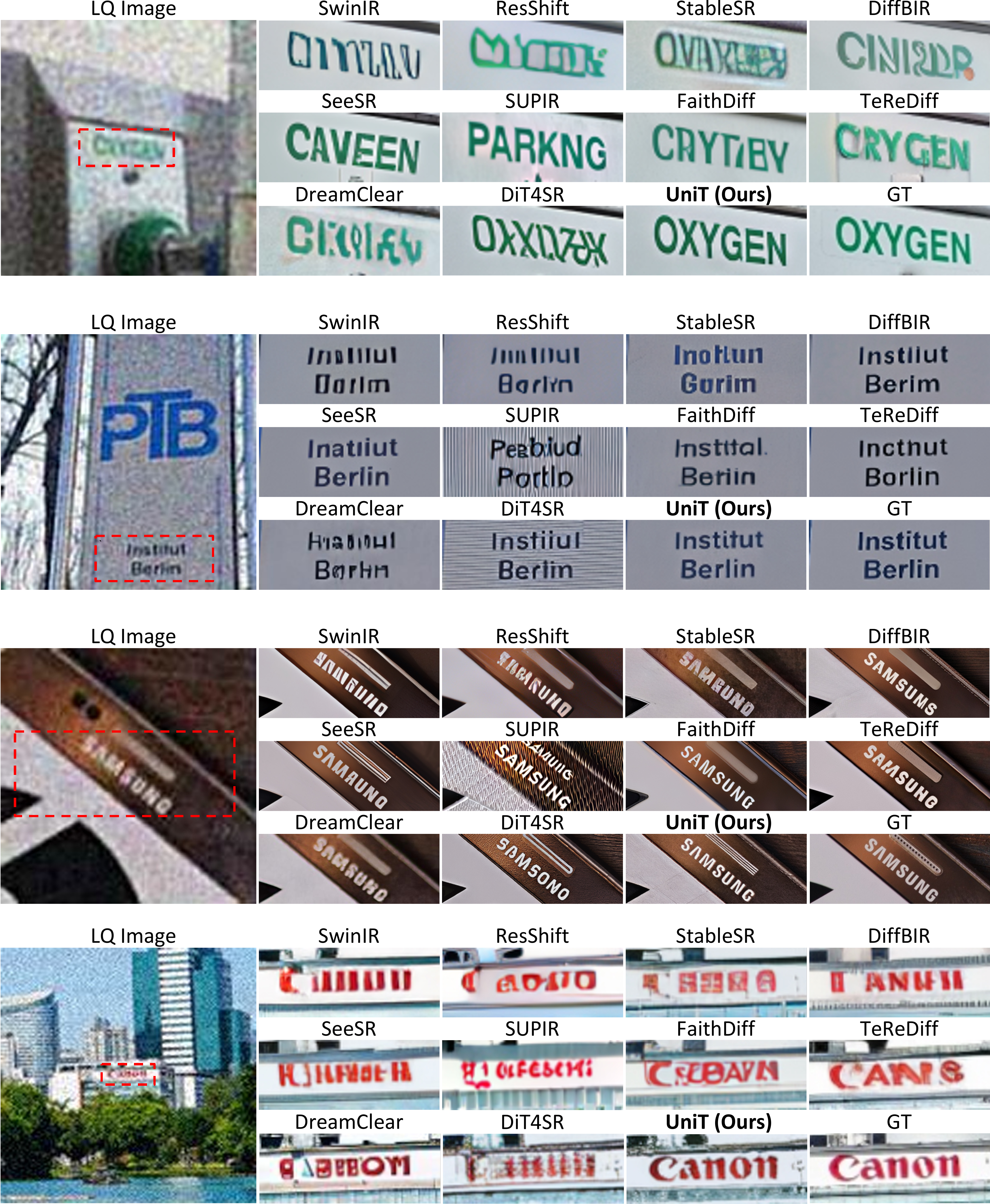}
    \caption{\textbf{Text restoration results on SA-Text test set Level 3.}}
    \label{fig:qual_satext_3}
\end{figure*}

Additional qualitative text restoration results for both the Real-Text and SA-Text benchmarks~\cite{min2025text} are presented in Fig.~\ref{fig:qual_realtext}, Fig.~\ref{fig:qual_satext_1}, Fig.~\ref{fig:qual_satext_2}, and Fig.~\ref{fig:qual_satext_3}.

\section{Limitation and Future Work}
\label{supple:sec_H_limitation}

Our UniT text restoration framework consists of three core components: a VLM, a TSM, and a DiT backbone. Although the TSM’s intermediate OCR predictions stimulate VLM self-correction and improve text restoration performance (Tab.~\ref{tab:abl_vlm_tsm}), we find that training a more robust or architecturally refined TSM could further strengthen VLM-assisted correction. Also, while the VLM can extract multilingual degraded text, the TSM is trained only on English, restricting self-correction to English predictions. Extending the TSM to multilingual OCR is a promising direction, especially when reliance on large, computationally intensive VLMs is undesirable. However, constructing a multilingual TSM is challenging due to diverse glyph structures, stroke patterns, and writing styles that vary significantly across languages and degrade unpredictably. Addressing these limitations either through the development of a multilingual TSM or through improved VLM mechanisms for extracting degraded text directly from low-quality images, offers a significant opportunity to enhance the robustness and expand the applicability of UniT in multilingual text restoration scenarios.


\clearpage
\newpage


{
    \small
    \bibliographystyle{ieeenat_fullname}
    \bibliography{main}
}

\end{document}